\documentclass[10pt,journal,compsoc]{IEEEtran}
\ifCLASSOPTIONcompsoc
\usepackage[nocompress]{cite}
\else
\usepackage{cite}
\fi
\ifCLASSINFOpdf
\else
\fi

\usepackage{amssymb}

\usepackage{amsfonts}

\usepackage{bbm}
\usepackage{bm}
\usepackage{color}
\usepackage{cite}
\usepackage[linesnumbered,ruled]{algorithm2e}
\usepackage{ulem} 
\usepackage{hyperref}
\usepackage{float}
\usepackage[T1]{fontenc}
\usepackage{booktabs}
\usepackage{multirow}
\usepackage{mathrsfs}
\usepackage[utf8]{inputenc}
\usepackage{ulem}
\usepackage{color}
\usepackage[table]{xcolor}

\newcounter{RNum}

\usepackage{graphics} 
\usepackage{epsfig} 
\usepackage{times} 
\usepackage{amsmath} 

\usepackage{amsfonts}

\usepackage{graphicx}
\usepackage{pifont}
\usepackage{color}
\usepackage{hyperref}
\usepackage{soul} 
\DeclareSymbolFont{largesymbol}{OMX}{yhex}{m}{n}
\DeclareMathAccent{\Widehat}{\mathord}{largesymbol}{"62}

\usepackage{graphics} 
\usepackage{epsfig} 
\usepackage{times} 
\usepackage{amsmath} 

\newcommand{\NoOne}[1]{\textcolor{red}{#1}}

\usepackage[caption=false,font=footnotesize]{subfig}

\usepackage{amsfonts}

\usepackage{graphicx}
\usepackage{pifont}

\soulregister\ref7
\soulregister\cite7
\soulregister\url7

\usepackage{booktabs}
\usepackage{multirow}
\usepackage{pifont}

\usepackage{ragged2e}
\usepackage{bbding}

\begin{document}
	%
	\title{Towards Real-World Visual Tracking with Temporal Contexts}
	%
	%
	%
	%
	
	\author{Ziang Cao,~\IEEEmembership{Student Member,~IEEE,}
		Ziyuan~Huang,~\IEEEmembership{Member,~IEEE,}
		Liang~Pan,~\IEEEmembership{Member,~IEEE,}
		Shiwei~Zhang,~\IEEEmembership{Member,~IEEE,}
		Ziwei~Liu,~\IEEEmembership{Member,~IEEE,}
		and~Changhong~Fu,~\IEEEmembership{Member,~IEEE}
		\IEEEcompsocitemizethanks{\IEEEcompsocthanksitem Z. Cao is with the School of Computer Science and Engineering, Nanyang Technological University, Singapore. \protect\\
			E-mail: ziang.cao@ntu.edu.sg
			\IEEEcompsocthanksitem Z. Huang is with the Department of Mechanical Engineering, National University of Singapore, Singapore. \protect\\
			E-mail: ziyuan.huang@u.nus.edu
			
			\IEEEcompsocthanksitem S. Zhang is with DAMO Academy, Alibaba Group. \protect\\
			E-mail: {zhangjin.zsw@alibaba-inc.com}
			
			\IEEEcompsocthanksitem L. Pan and Z. Liu (Corresponding author) are with the School of Computer Science and Engineering, Nanyang Technological University, Singapore. \protect\\
			E-mail: {liang.pan, ziwei.liu}@ntu.edu.sg
			
			\IEEEcompsocthanksitem C. Fu (Corresponding author) is with the School of Mechanical Engineering, Tongji University, Shanghai 201804, China. \protect\\
			E-mail: changhongfu@tongji.edu.cn
			
		}
		
		\thanks{Manuscript received July 27, 2022.}
	}
	
	%
	%

	\markboth{IEEE TRANSACTION ON PATTERN ANALYSIS AND MACHINE INTELLGENCE}%
	{Shell \MakeLowercase{\textit{et al.}}: Bare Demo of IEEEtran.cls for IEEE Journals}
	%



	\IEEEtitleabstractindextext{%
		\begin{abstract}
			\justifying
			Visual tracking has made significant improvements in the past few decades. Most existing state-of-the-art trackers 1) merely aim for performance in ideal conditions while overlooking the real-world conditions; 2) adopt the tracking-by-detection paradigm, neglecting rich temporal contexts; 3) only integrate the temporal information into the template, where temporal contexts among consecutive frames are far from being fully utilized. To handle those problems, we propose a two-level framework (TCTrack) that can exploit temporal contexts efficiently. Based on it, we propose a stronger version for real-world visual tracking, \textit{i.e.}, TCTrack++. It boils down to two levels: \textbf{features} and \textbf{similarity maps}. Specifically, for feature extraction, we propose an attention-based temporally adaptive convolution to enhance the spatial features using temporal information, which is achieved by dynamically calibrating the convolution weights. For similarity map refinement, we introduce an adaptive temporal transformer to encode the temporal knowledge efficiently and decode it for the accurate refinement of the similarity map. To further improve the performance, we additionally introduce a curriculum learning strategy. Also, we adopt online evaluation to measure performance in real-world conditions. Exhaustive experiments on 8 well-known benchmarks demonstrate the superiority of TCTrack++. Real-world tests directly verify that TCTrack++ can be readily used in real-world applications. 

		\end{abstract}
		
		}

	\maketitle

	\IEEEdisplaynontitleabstractindextext

	%
	\IEEEpeerreviewmaketitle

	\IEEEraisesectionheading{\section{Introduction}\label{sec:introduction}}

	%
	%
	%
	%
	\IEEEPARstart{A}{s} one of the most fundamental tasks in computer vision, visual tracking has received much attention and has many applications. Recently, tracking-by-detection framework has achieved state-of-the-art (SOTA) performance in visual tracking~\cite{siamrpn,siamrpn++,siamcar,siamban}. Despite impressive accuracy and robustness, most existing trackers neglect the rich temporal information in tracking scenarios, thereby highly limiting the rise of tracking performance, especially in scenes with severe motion.

	To further boost the tracking performance, some works have been devoted to exploiting the temporal contexts. For discriminative correlation filter (DCF)-based methods, the variation in the response maps along the temporal dimension is penalized~\cite{Huang2019ICCV,Li_2020_CVPR}, which guides the current response map by previous ones. However, as a result of poor representative feature expression without using large-scale data, the tracking performance is limited. In Siamese networks, which is the focus of this paper, temporal information is introduced in most works through dynamic templates. By integrating historical information into the current template, dynamic templates can be formulated, \textit{e.g.}, through weighted sum~\cite{updatenet}, graph network~\cite{gct}, transformer structure~\cite{yan2021learning,trdimp}, and memory networks~\cite{stmtrack,yang2018learning}. In spite of those pioneering works in introducing temporal information into tracking tasks, most of them merely consider \textit{only a single level}, \textit{i.e.}, aggregate template information via their proposed special module. The consecutive information from feature extraction and similarity maps is still overlooked. Moreover, since most temporal trackers choose to store historical features to achieve SOTA performance, their high-memory consumption and complex structure significantly impede the tracking performance in real-world conditions, especially in applications on resource-limited platforms.

\begin{figure}[t]
		\centering

		\includegraphics[width=0.47\textwidth]{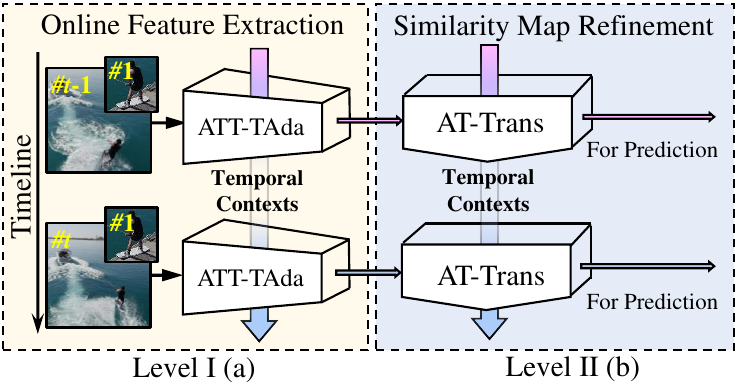}
		
		\caption{
			Overview of our framework. It exploits temporal information efficiently via two levels: (a) \textit{feature level} by the attention-based temporally adaptive convolutional neural networks (ATT-TAdaCNN) and (b) \textit{similarity maps level} by the adaptive temporal transformer (AT-Trans). 
		}
		\label{fig:page1}
		
	\end{figure}

	To address those issues mentioned in the existing trackers, this work presents a two-level efficient framework to introduce temporal contexts for real-world visual tracking, namely TCTrack++. As shown in Fig.~\ref{fig:page1}, TCTrack++ introduces temporal context into the tracking pipeline \textbf{via} \textbf{two} \textbf{levels}, \textit{i.e.,} features and similarity maps. 
	At the \textbf{feature} \textbf{level}, compared with the standard spatial-only backbone adopted by most existing temporal trackers, we are the first to propose attention-based temporally adaptive convolution (ATT-TAdaConv) for exploring temporal contexts during feature extraction. Specifically, instead of directly storing historical frames, we integrate the historical search features into a fixed-size small dynamic calibration vector via a cross-attention operation. Based on the vector, the extracted CNN features will involve rich temporal information. Therefore, compared with TCTrack, 1) we avoid the hyper-parameters associated with historical frames; 2) regardless of the length of tracking scenarios, the memory loading and computation complexity of our tracker is constant and low.
	At the \textbf{similarity} \textbf{map} \textbf{level}, an adaptive temporal transformer (AT-Trans) is proposed to refine the similarity map according to the temporal information. Specifically, AT-Trans adopts an encoder-decoder structure, where \textbf{(i)} the encoder produces the temporal prior knowledge for the current time step, by integrating the previous prior with the current similarity map, and \textbf{(ii)} the decoder refines the similarity map based on the produced temporal prior knowledge in an adaptive way. 
	Different from \cite{stmtrack,trdimp,gct}, we defined a fix-sized temporal prior knowledge that keeps updating at each frame. Therefore, our AT-Trans is memory efficient and edge-platform friendly. Moreover, we adopt a curriculum learning strategy, training our tracker from easier data to harder data, \textit{i.e.}, from a small temporal range to a large one. Overall, these novel methods allow our temporal framework TCTrack++ to exploit temporal contexts comprehensively while maintaining efficiency. Extensive offline and online evaluations on 8 well-known benchmarks strongly demonstrate the superior effectiveness and the efficiency of our proposed framework. To further verify the tracking performance in real-world tracking scenes, we conduct the real-world tests in many scenes including indoor and outdoor scenes on a typical resource-limited platform, \textit{i.e.}, NVIDIA Jetson AGX Xavier. The results show that our TCTrack++ achieves impressive tracking performance with a real-time speed.
	


	This paper is an extension of our conference version accepted to CVPR2022~\cite{cao2022tctrack}. Compared with the earlier version, \textit{i.e.}, TCTrack, we provide novel and valuable materials including new architectures, exhaustive experiments, a comprehensive survey, and detailed implementations which are clearly summarized as follows:
	
	\begin{itemize}
	
		\item To further explore temporal contexts from tracking sequences, we introduce the attention mechanism and build ATT-TAdaConv. Owing to the no-prior design and lightweight architecture, our TCTrack++ is competent for modeling temporal contexts effectively and robustly.
		
		\item To lower the memory load during training and achieve better convergence, we adopt a curriculum learning strategy by transforming from easier data (small temporal range) to harder data (large temporal range).

		\item 	We conduct comprehensive evaluations on 8 benchmarks which can be divided into two parts: 1) online evaluation to measure the real-world tracking performance more realistically; 2) offline evaluation to evaluate tracking performance regardless of inference time and latency.

		\item We also implement our tracker on many real-world tracking indoor and outdoor scenes which require low latency and real-time speed. The tests comprehensively validate the promising performance of our framework.

		\item On top of the conference version, we provide more analysis of existing trackers using temporal information and more detailed information about implementation and our proposed structure.
	
	\end{itemize}
	
	\section{Related Work}\label{sec:Related Work}
	
	\begin{figure*}[t]
		\centering

		\includegraphics[width=1\textwidth]{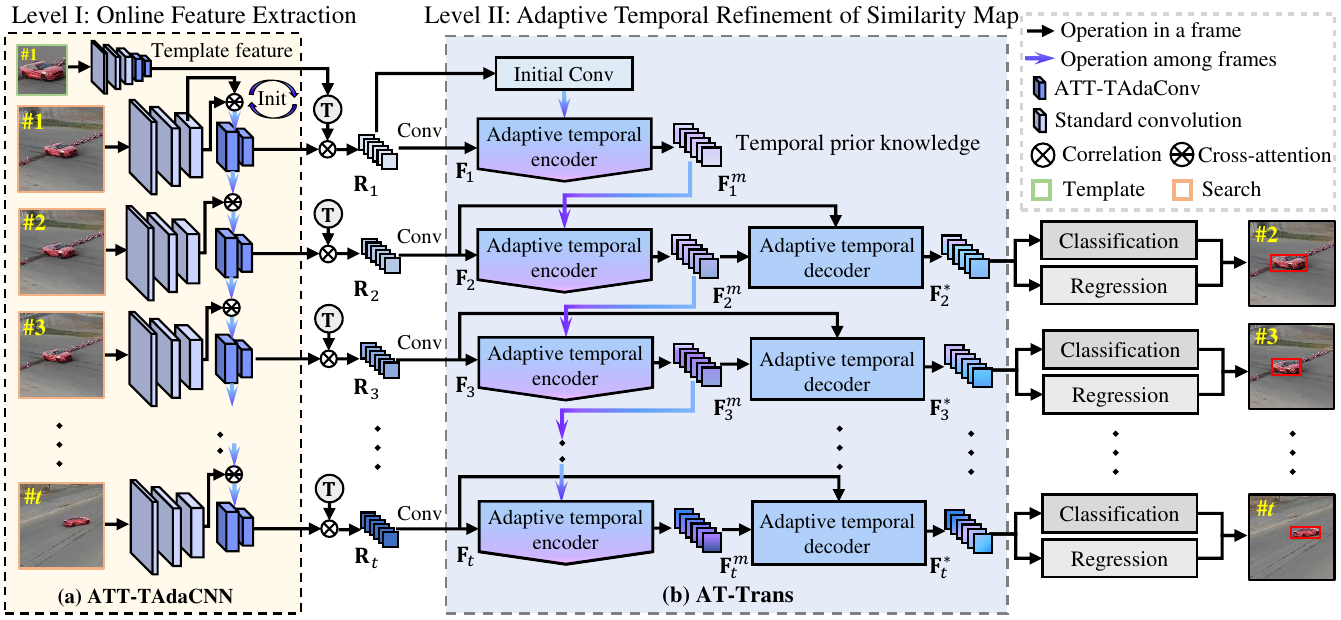}
		\vspace{-20pt}
		\caption{Overview of our two-level tracking framework. It mainly consists of three components, \textit{i.e.}, ATT-TAdaCNN for online feature extraction shown in Fig.~\ref{fig:conv}, AT-Trans for similarity map refinement shown in Fig.~\ref{fig:work}, and classification\&regression for final prediction. This figure shows the workflow of our TCTrack++ when tracking sequences are \textit{t} frames. Comprehensive temporal knowledge is introduced in our framework via two levels. Best view in color.}
		\vspace{-10pt}
		\label{fig:1}
	\end{figure*}
	
	\subsection{Tracking by Detection Methods}

	After D. S. Bolme \textit{et al.} firstly proposed the MOSSE filter~\cite{Bolme2010CVPR}, there are many works trying to boost the tracking performance of DCF-based trackers~\cite{kcf,SRDCF,strcf,li2021adtrack}. However, suffering from low representative features, the improvement of tracking performance is highly limited. In recent, tracking by detection has been one of the most common pipelines in the Siamese-based trackers. Starting from SiamFC~\cite{siamfc}, SiamRPN~\cite{siamrpn} and SiamRPN++~\cite{siamrpn++} improve the tracking paradigm via introducing region proposal network (RPN)~\cite{ren2015faster}. Then, many valuable works have been made to improve the tracking performance~\cite{siamdw,siamban,siamcar,siamgat}. Moreover, to meet the requirement of aerial conditions, some works~\cite{siamapn++,siamapn,hift} try to keep the promising trade-off between performance and speed. They can maintain competitive performance with SOTA trackers while maintaining real-time speed on recourse-limited aerial platform.

	Despite achieving SOTA performance, those trackers above disregard the temporal contexts in the tracking scenarios, thereby blocking the performance improvement. Differently, our tracker can effectively model the historical temporal contexts during the tracking for increasing the discriminability and robustness.

	\subsection{Temporal-Based Tracking Methods}
	
	%
	
	Actually, some works have been devoted to exploiting the temporal information in tracking scenarios for raising the tracking performance. For DCF-based trackers, they explore temporal regularized methods to exploit historical frames effectively~\cite{Huang2019ICCV,Li_2020_CVPR,yan2021learning}. However, their performance is still not competitive with those SOTA Siamses trackers. As for deep learning-based methods, most existing methods aim to build an effective dynamic template. Based on the adjusted template, those trackers can exploit the temporal information for tracking. UpdateNet~\cite{yan2021learning} combines the two historical templates and current ones to encode the temporal information. Similarly, STARK~\cite{yan2021learning} explicitly updates the dynamic template and utilizes a transformer module to exploit temporal information. Recently, ARTrack~\cite{wei2023autoregressive} adopts a autoregressive framework for propagating preceding motion dynamic into succeeding frames. Despite the superior performance, they only encode the discrete historical frames into tracking, which loses valuable information. To explore temporal contexts from consecutive frames clips, transformer integration~\cite{trdimp}, template memory update~\cite{yang2018learning,stmtrack}, and graph network~\cite{gct} are conducted via defining key hyper-parameters associated with the length of clips. Adopting consecutive frames enriches the temporal contexts in the template feature and then promotes tracking performance. However, the adjustment of hyper-parameters is time-consuming and has a significant influence in terms of accuracy and speed.

	Furthermore, most existing temporal-based trackers focus on a single feature level, \textit{i.e.}, template features. Besides, the backbone of those trackers for feature extraction is merely spatial-aware. The temporal information is not fully utilized. Meantime, the real-world application of most trackers also suffer from heavy structure and high memory loading. Therefore, in this paper, we build a comprehensive and efficient temporal framework via two levels and a low memory loading structure. It not only exploits the consecutive temporal contexts from the whole tracking scenarios rather than video clips online but also avoids the increasing memory loading and hyper-parameters.

	\subsection{Temporal Modelling in Videos}

	Modelling the temporal dynamics is essential for a genuine understanding of videos. Hence, it is widely explored in both supervised~\cite{feichtenhofer2019slowfast,tran2018r2p1d,huang2021tadaconv,liu2021tam,wang2018nonlocal,lin2019tsm} and self-supervised paradigms~\cite{huang2021mosi,han2019dpc,kim2019stpuzzle,huo2021csj,han2020memdpc,qing2022hico,qian2021cvrl}.
	In self-supervised learning, a video model learns spatio-temporal representations by solving various manually designed pre-text tasks, such as dense future prediction~\cite{han2019dpc,han2020memdpc}, jigsaw puzzle solving~\cite{kim2019stpuzzle,huo2021csj}, and pseudo motion classification~\cite{huang2021mosi}, \textit{etc.}
	On top of the instance discrimination task~\cite{wu2018instancediscrimination}, the contrastive learning framework is formed~\cite{chen2020simclr,he2020moco} and promoted to video applications~\cite{qing2022hico,qian2021cvrl}.
	Supervised video recognition explores various connections between different frames, such as 3D convolutions~\cite{tran2015c3d}, temporal convolutions~\cite{tran2018r2p1d}, temporal shift~\cite{lin2019tsm}, spatio-temporal relations~\cite{wang2018nonlocal}, \textit{etc.}
	Closely related to our work is the temporally adaptive convolutions~\cite{huang2021tadaconv}, which serves as a plug-and-play module for video models that adaptively adjusts the convolution weights of each frame according to its local and global temporal contexts. In this work, to adapt to the tracking task, we propose an online CNN which can extract spatial features according to temporal contexts for enriching the temporal information comprehensively.

	\subsection{Online Evaluations}\label{onlineeva}
	
	In the visual tracking field, most existing evaluations are offline. They aim to split each frame and evaluate the tracking performance independently~\cite{otb,huang2019got}. Recently, M. Li \textit{et al.}~\cite{li2020towards} propose an online evaluation method that is more realistic for real-world tasks and B. Li \textit{et al.}~\cite{li2021predictive} extends it to the visual tracking field, who considers both the influence of speed and accuracy on tracking performance.
	
	Different from the ideal tracking condition, in real-world conditions, when the tracker completes the tracking process, \textit{i.e.}, receives the $t$-th frame and gets the results of $t$-th frame, the object has moved away during the inference time. Therefore, only when a tracker is fast and effective, the performance of real-world online evaluation will high which is correspond to our intuition. In this paper, we adopt both offline evaluation and online evaluation for exhaustively evaluating the real-world tracking performance of our tracker.



	
	\section{Temporal Contexts for Real-World Tracking}\label{sec:Proposed Method}
	
	In this section, we will introduce the detailed structure and workflow of our framework as shown in Fig.~\ref{fig:1}. The proposed method consist of two aspects: \textbf{(1)} online feature extraction achieved by ATT-TAdaCNN (Sec.~\ref{feat}) which can incorporate the temporal contexts into spatial CNN features without increasing much computation; \textbf{(2)} refinement of similarity maps where we propose a novel AT-Trans to encode the temporal knowledge from consecutive scenarios and then refine the similarity map according to the temporal prior knowledge (Sec.~\ref{tf}).

	\begin{figure*}[t]
		\centering

		\includegraphics[width=1\textwidth]{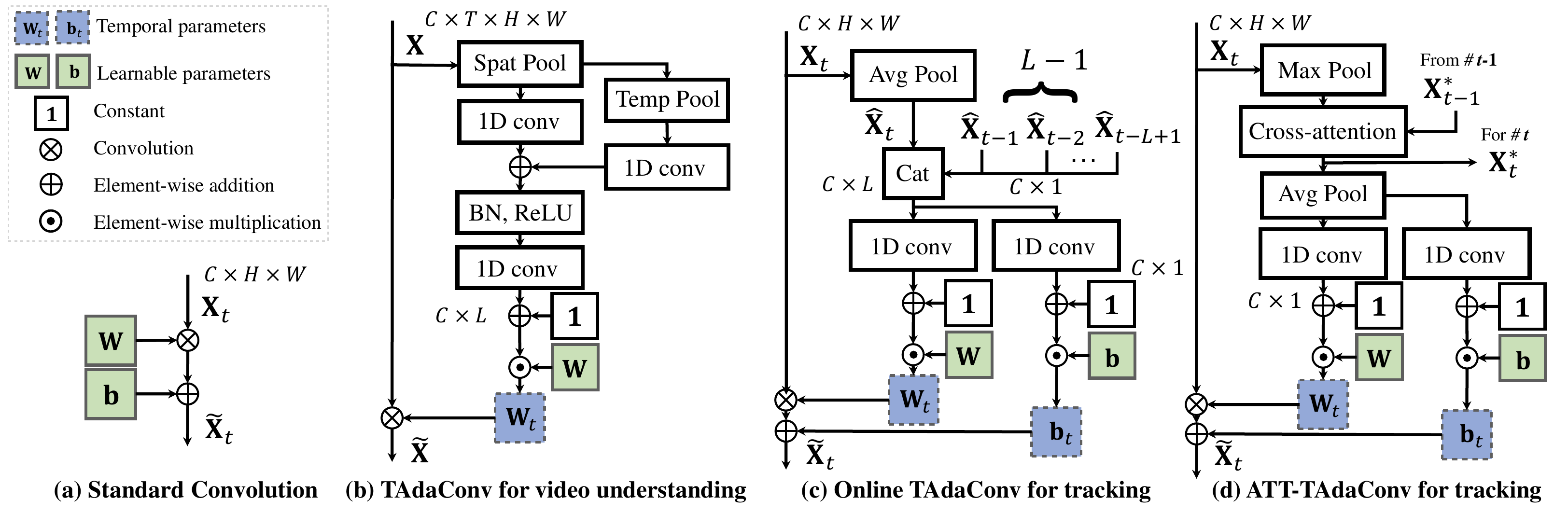}
		
		\caption{The workflow of our ATT-TAdaConv (d). Different from Online TAdaConv (c), the temporal context is integrated via a cross-attention module. Based on this strategy, our ATT-TAdaConv can avoid the hyper-parameters (L in Online TAdaConv (c)) and aggregate the temporal information adaptively.}
		\label{fig:conv}
		
	\end{figure*}
	
	\noindent\textbf{Revisit Multi-Head Attention.} 
	As a fundamental component of the transformer~\cite{aaat}, multi-head attention is formulated as follows:
	\vspace{-10pt}
	\begin{equation}\label{e1}
		\begin{aligned}
			&\mathrm{MultiHead}(\mathbf{Q},\mathbf{K},\mathbf{V})=\Big(\mathrm{Cat}(\textbf{H}_{att}^1,...,\textbf{H}_{att}^N)\Big)\mathbf{W}\\
			&\textbf{H}_{att}^{n}=\mathrm{Attention}(\mathbf{Q}\mathbf{W}^n_q,\mathbf{K}\mathbf{W}_k^n,\mathbf{V}\mathbf{W}_v^n)\\
			&\mathrm{Attention}(\mathbf{Q},\mathbf{K},\mathbf{V})=\mathrm{Softmax}(\mathbf{Q}\mathbf{K}^{\mathrm{T}}/\sqrt{d})\mathbf{V}\\
		\end{aligned}
		~ ,
	\end{equation}
	where $\mathbf{Q}, \mathbf{K},$ and $\mathbf{V}$ represent the query, key, and value in the attention operation, $\mathbf{W}\in~\mathbb{R}^{C_i\times C_i}$, $\mathbf{W}^n_q\in~\mathbb{R}^{C_i\times C_h}$, $\mathbf{W}^n_k\in~\mathbb{R}^{C_i\times C_h}$, and $\mathbf{W}^n_v\in~\mathbb{R}^{C_i\times C_h}$ are learnable weights, and $\sqrt{d}$ is the scaling factor to avoid gradient vanishing.

	\subsection{Online Feature Extraction with ATT-TAdaConv}
	\label{feat}

	Compared with other temporal trackers, we are the first to introduce the temporal backbone for visual tracking tasks. It can incorporate the spatial CNN features with temporal contexts. The comparison among standard convolution, TAdaConv~\cite{huang2021tadaconv}
	, Online TAdaConv~\cite{cao2022tctrack}, and ATT-TAdaConv are illustrated in Fig.~\ref{fig:conv}. Based on our previous version (Online-TAdaConv), we propose a novel attention-based TAdaConv for considering temporal contexts. By using the fix-sized temporal knowledge strategy similar to AT-Trans, it eliminates the hyper-parameters associated with the video length ($L$) and adopts a more effective attention-based way to explore global interdependencies between historical and current features.  
	
	For clarification, we denote the input features of the $t$-th frame as $\mathbf{X}_t \in \mathbb{R}^{C \times H \times W}$. The output of the ATT-TAdaConv $\mathbf{\tilde{X}}_t\in \mathbb{R}^{C \times H \times W}$ can be formulated as follows:
	\begin{equation}
		\mathbf{\tilde{X}}_t = \mathbf{W}_t * \mathbf{X}_t + \mathbf{b}_t\ ,
	\end{equation}
	\noindent where the operator $*$ denotes the convolution operation and $\mathbf{W}_t, \mathbf{b}_t$ are the weight and bias of ATT-TAdaConv. In contrast to a standard convolution layer where the weight and bias are learnable parameters (constant during tracking), our ATT-TAdaConv generates calibration factors to adjust the learnable parameters $\mathbf{W}_t, \mathbf{b}_t$ for each frame according to the temporal contexts as follows:
	
	\begin{equation}
		\mathbf{W}_t=\mathbf{W}_b\cdot\bm{\alpha}^w_t,~
		\mathbf{b}_t=\mathbf{b}_b\cdot\bm{\alpha}^b_t~,
	\end{equation}
	where $\bm{\alpha}^w_t$ and $\bm{\alpha}^b_t$ represent the calibration factors. Given the small size of calibration factors, our module can integrate temporal information and maintain promising efficiency.

	Besides, different from the original structure in video understanding, our ATT-TAdaConv merely utilizes the temporal contexts in the past. To avoid the hyper-parameter associated with the temporal context queue, we introduce the fixed-size temporal knowledge $\mathbf{X}_t^*$ to accumulate the historical features. To lower the calculation, we adopt adaptive max pooling to decrease the size of features and a convolutional layer to lower the number of channels. Therefore, the cross-attention operation can be formulated as follows:

	\begin{equation}
		\begin{aligned}
			&\mathbf{X_t^*} =  \mathrm{Attention}(\mathcal{F}_q(\mathbf{X_{t}'}),\mathcal{F}_k(\mathbf{X_{t-1}^*}),\mathcal{F}_v(\mathbf{X_{t-1}^*}))\\
			&~~~~~~~~~~~~~~~~~~~~~~~\mathbf{X_{t}'}=\mathrm{MaxPool}(\mathbf{X_{t}})\\
		\end{aligned}
		~ ,
	\end{equation}
	where $\mathcal{F}_i, i=q,k,v$ represents the convolutional operation.
	
	For efficiency, we adopt the adaptive average pooling on the temporal knowledge $\mathbf{X_{t}}$. Besides, to ensure the effectiveness of our method, we introduce the residual connection in the generation of the calibration factor. Thus, the calibration factors $\bm{\alpha}^w_t$ and $\bm{\alpha}^b_t$ can be calculated by:
	
	\begin{equation}
		\begin{aligned}
			&\mathbf{\hat{X}}_t=\text{GAP}(\mathbf{X}_t^*)\\
			&\bm{\alpha}^w_t =\mathcal{F}_\text{w}(\mathbf{\hat{X}}) + 1\\
			&\bm{\alpha}^b_t=\mathcal{F}_\text{b}(\mathbf{\hat{X}}) + 1\\
		\end{aligned}
		~ ,
	\end{equation}
	where the GAP represents the global average pooling.
	
	Please note that the learnable weights of $\mathcal{F}_w$ and $\mathcal{F}_b$ are initialized to zero so that $\mathbf{W}_t=\mathbf{W}_b$ and $\mathbf{b}_t=\mathbf{b}_b$ before training.

	For $t=1$, where there is not enough historical knowledge, we adopt a convolutional operation to initialize $\mathbf{X_{0}^*}$ as follows:
	\begin{equation}
		\mathbf{X_{0}^*}=\mathcal{F}_{init}(\mathbf{X_{1}'})~.
	\end{equation}

	Given our backbone $\varphi_{tada}$ that considers the temporal contexts in the feature extraction process, the similarity map $\textbf{F}_t$ for the $t$-th frame can be obtained as:
	
	\begin{equation}\label{11}
		\begin{aligned}
			&\textbf{R}_t=\varphi_{tada}(\textbf{Z})\star\varphi_{tada}(\textbf{X}_t)\\
			&\textbf{F}_t=\mathcal{F}(\textbf{R}_t)\\
		\end{aligned}
		~ ,
	\end{equation}
	
	\noindent where $\mathbf{Z}$ denotes the template and $\star$ represents the depth-wise correlation~\cite{siamrpn++}. 
	
	To the best of our knowledge, our method is the first attempt to efficiently integrate temporal information in the spatial CNN feature extraction network during tracking.

	\subsection{Similarity Refinement with AT-Trans}\label{tf}
	To enrich the temporal contexts in similarity maps, in this work, we also propose an AT-Trans for refining the similarity map $\mathbf{F}_t$.  Benefiting from the encoder-decoder structure, our AT-Trans can complete the temporal knowledge during encoding and refine the similarity maps according to the completed temporal knowledge during decoding. Note that in our AT-Trans, we employ multi-head attention with 6 heads, \textit{i.e.,} $N=6$ and $C_h$=$C_i/6$.

	Compared to CNN, Transformer achieve promising performance in encoding global context information effectively~\cite{aaat,dosovitskiy2021an}. Hence, it is an ideal framework to exploit the global temporal contexts. However, most existing temporal-based methods generally store the input features for temporal modeling, inevitably introducing sensitive parameters and unnecessary computation. To raise the efficiency, we eliminate the unnecessary operations and sensitive parameters by adopting online update strategy for temporal knowledge.

	


	
\begin{figure}[t]
		\centering
		\includegraphics[width=1.0\linewidth]{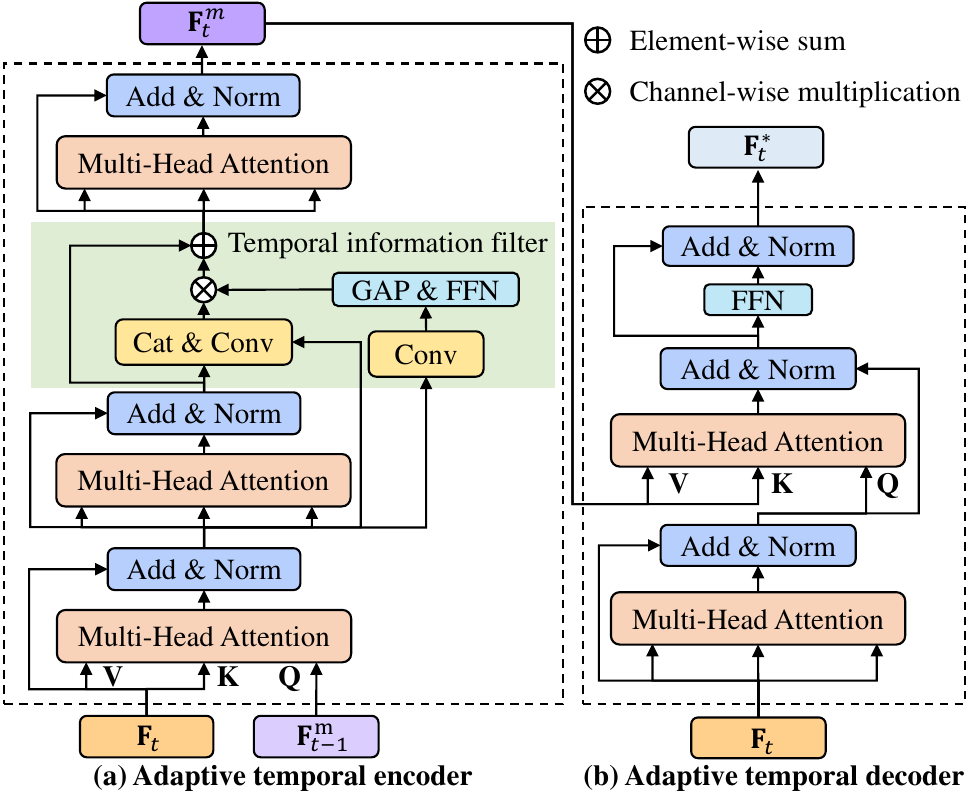}
		\vspace{-20pt}
		\caption{Structure of the adaptive temporal transformer. The left sub-window illustrates the adaptive temporal encoder to model the temporal knowledge. The right sub-window shows the component of the decoder. Best viewed in color.
		}
		\vspace{-10pt}
		\label{fig:work} 
	\end{figure}

	\noindent\textbf{Transformer encoder.} The function of the encoder is to accumulate temporal knowledge from previous features and current ones.
	We stack two multi-head attention layers before a temporal information filter to explore the interdependencies between features. 
	After filtering noise information, the final temporal prior knowledge for the current step is obtained by further attaching a multi-head attention layer to the filtered information. 
	The structure of the encoder is presented in Fig.~\ref{fig:work}(a).
	
	Given the previous temporal prior knowledge $\mathbf{F}_{t-1}^m$ and the current similarity map $\mathbf{F}_{t}$, there are two ways to integrate their information into the current prior knowledge $\mathbf{F}_{t}^m$, with respect to the selection of the query, key, and values. 
	One uses $\mathbf{F}_{t-1}^m$ as the query and $\mathbf{F}_{t}$ as the value and key, while the other uses them in reverse.
	In our method, we adopt the former, as this essentially puts more emphasis on the current similarity map. 
	It is obviously that closer temporal information is more valuable than the previous one for representing the characteristics of the current object more accurately. Thus, adopting $\mathbf{F}_{t}$ as the values will introduce more valuable prior knowledge.
	Empirical results in Sec.~\ref{Sec:abla} also validate the effectiveness of this choice. 
	Hence, we obtain the output of the stacked multi-head attention layer in $t$-th frame $\mathbf{F}^{2}_t$ by:
	\begin{equation}
		\begin{aligned}
			& \mathbf{F}^{1}_t=\mathrm{Norm}(\mathbf{F}_{t}+\mathrm{MultiHead}(\mathbf{F}_{t-1}^m,\mathbf{F}_{t},\mathbf{F}_{t})) \\
			& \mathbf{F}^{2}_t=\mathrm{Norm}(\mathbf{F}^{1}_{t}+\mathrm{MultiHead}(\mathbf{F}^{1}_{t},\mathbf{F}^{1}_{t},\mathbf{F}^{1}_{t}))
		\end{aligned} \ ,
	\end{equation}
	where $\mathrm{Norm}$ represents the layer normalization. 

	Actually, the temporal information is incorporated in $\mathbf{F}^{2}_t$, which can be used to guide similarity refinement. Nevertheless, the temporal information will mislead the tracking sometimes when encountering motion blur or occlusion.
	In those conditions, some unwanted contexts may be included if we pass along the complete temporal information without any filtering. To eliminate the unwanted information, a neat temporal information filter is proposed by attaching a feed-forward network $\mathrm{FFN}$ to the global descriptor of $\mathbf{F}^{1}_t$ obtained by global average pooling $(\mathrm{GAP})$, \textit{i.e.,} $\bm{\alpha}=\mathrm{FFN}(\mathrm{GAP}(\mathcal{F}(\mathbf{F}^{1}_t)))$. The filtered information $\mathbf{F}^{f}_t$ is obtained by:
	\begin{equation}
		\mathbf{F}^{f}_t=\mathbf{F}^{2}_t+\mathcal{F}(\mathrm{Cat}(\mathbf{F}^{2}_t,\mathbf{F}^{1}_t))*\bm{\alpha}
		~ ,
	\end{equation}
	where $\mathcal{F}$ denotes a convolution layer. 
	With this, the temporal knowledge of \textit{t}-th frame, $\mathbf{F}^{m}_t$ can be obtained as follows:
	\begin{equation}
		\mathbf{F}^{m}_t=\mathrm{Norm}(\mathbf{F}^{f}_t+\mathrm{MultiHead}(\mathbf{F}^{f}_t,\mathbf{F}^{f}_t,\mathbf{F}^{f}_t))
		~ .
	\end{equation}
	
	Based on the online updating strategy and fixed temporal prior knowledge, we not only integrate the temporal information but makes TCTrack++ memory-efficient compared to approaches that require saving all the intermediate temporal information. Consequently, our AT-Trans adaptively encodes the temporal prior in a memory-efficient way.
	
	Please not that for the first frame in a tracking sequence, we adopt a special initialization method. It is unreasonable that using a unified initialization for the initial temporal prior $\mathbf{F}_{0}^m$ when tracking different object. Besides, observing that the similarity map in the first frame essentially represents the semantic features of the target object, thereby we set the initial temporal prior by a convolution over the initial similarity map $\mathbf{F}_{0}$, \textit{i.e.,} $\mathbf{F}_{0}^m=\mathcal{F}_{init}(\mathbf{R}_{1})$. We also empirically show our initialization is better in Sec.~\ref{Sec:abla}.



	\noindent\textbf{Transformer decoder.} 
	According to the temporal prior knowledge $\mathbf{F}^{m}_t$, the decoder can refine the similarity map. 
	Note that to better explore the interrelations between temporal knowledge and current spatial features $\mathbf{F}_t$, we adopt two multi-head attention layers with feed-forward before output. 
	Its structure is presented in Fig.~\ref{fig:work}(b).
	By generating the attention map, the valid information in the temporal knowledge $\mathbf{F}^{m}_t$ can be extracted for refining the similarity map $\mathbf{F}_t$ to obtain the final output $\mathbf{F}^*_t$:
	\begin{equation}
		\begin{split}
			&\mathbf{F}^3_t=\mathrm{Norm}(\mathbf{F}_t+\mathrm{MultiHead}(\mathbf{F}_t,\mathbf{F}_t,\mathbf{F}_t))\\
			&\mathbf{F}^4_t=\mathrm{Norm}(\mathbf{F}^3_t+\mathrm{MultiHead}(\mathbf{F}^3_t,\mathbf{F}^m_t,\mathbf{F}^m_t))\\
			&\mathbf{F}^{*}_t=\mathrm{Norm}(\mathbf{F}^{4}_t+\mathrm{FFN}(\mathbf{F}^{4}_t))\\
		\end{split}
		~ .
	\end{equation}
	
	Relying on the encoder-decoder structure of AT-Trans, the temporal contexts are effectively exploited to refine the similarity maps for boosting robustness and accuracy. To the best of our knowledge, AT-Trans is the first attempt to explore temporal contexts from similarity maps for visual tracking.

%
%
%
%
%

		\subsection{Loss Functions}\label{loss}
	After obtaining the refined features $\mathbf{F}^{*}_t$, the tracking results can be calculated via a standard classification and regression network. Specifically, Given $\mathbf{F}^*_t$, the output of classification and regression branches can be obtained as follows:
	\begin{equation}
	\begin{split}
	&\mathbf{F}^{cls1}_t=\varphi_{cls1}(\mathbf{F}^*_t), ~~ \mathbf{F}^{cls2}_t=\varphi_{cls2}(\mathbf{F}^*_t)\\
	&~~~~~~~~~~~~~~~~~\mathbf{F}^{loc}_t=\varphi_{loc}(\mathbf{F}^*_t)
	\end{split}
	~ .
	\end{equation}
	To improve the precision of regression, we design a new branch named $\mathcal{L}_{loc1}$. For clarification, we denote the center and scale of ground truth bounding box as $\mathbf{G}^{loc}_t=(\mathbf{x},\mathbf{y},\mathbf{w},\mathbf{h})$. Therefore, the $\mathcal{L}_{loc1}$ and $ \mathcal{L}_{loc2}$ can be formulated as follows:
	\begin{equation}
		\begin{split}
			&\mathcal{L}_{loc1}=\dfrac{1}{4N^2}\sum_{i=1}^{4}\sum_{j=1}^{N}\sum_{l=1}^{N} \textbf{M}(i,j,l) *\mathbf{D}(i,j,l)\\
			&\mathbf{D}=\sqrt{\dfrac{(\mathbf{F}^{loc1}_t[0]-\mathbf{x})^2}{\mathbf{w}}+\dfrac{(\mathbf{F}^{loc1}_t[1]-\mathbf{y})^2}{\mathbf{h}}}\\
			&\mathcal{L}_{loc2}=\dfrac{1}{4N^2}\sum_{i=1}^{4}\sum_{j=1}^{N}\sum_{l=1}^{N}\textrm{IoU}(\mathbf{F}^{loc}_t,\mathbf{G}^{loc}_t)\\
		\end{split}
		~ ,
	\end{equation}
	where $\textbf{M} \in \mathbb{R}^{N \times N}$ is a mask, following~\cite{hift} and $\mathcal{L}_{loc2}$ represent the Intersection over Union (IoU) loss. .
	
	The part of classification consists of two branches, \textit{e.g.}, $\mathcal{L}_{cls1}$ and $\mathcal{L}_{cls2}$, adopting cross-entropy (CE) loss and binary cross-entropy (BCE) loss. Thus, the classification loss can be formulated as follow:
	\begin{equation}
	\begin{split}
	&\mathcal{L}_{cls1}=\mathcal{L}_{CE}(\mathbf{F}^{cls1}_t,\mathbf{G}^{cls1}_t)\\
	&\mathcal{L}_{cls2}=\mathcal{L}_{BCE}(\mathbf{F}^{cls2}_t,\mathbf{G}^{cls2}_t)\\
	\end{split}
	~ ,
	\end{equation}
	where $\mathbf{G}^{cls1}_t,~\mathbf{G}^{cls2}_t$ represent the ground truth for classification.
	Therefore, the overall loss function can be determined as follows:
	\begin{equation}
		\begin{split}
			&\mathcal{L}=\mathcal{L}_{cls1}+\mathcal{L}_{cls2}+\mathcal{L}_{loc1}+\mathcal{L}_{loc2}\\
		\end{split}
		~ .
	\end{equation}
	
	\begin{table}[t]
		\footnotesize
		\setlength{\tabcolsep}{4mm}
		\centering
		\caption{Comparison of inference time and parameters on NVIDIA Jetson AGX Xavier. Here, we use 287$\times$287$\times$3 as the input image and only evaluate the inference time of the CNN. 
		}
		
		\centering
		{
			\begin{tabular}{l|c|c }
				\toprule[2pt]
				
				{\textbf{Backbone}}&Inference time &Parameters\\
				\midrule
				{AlexNet~\cite{krizhevsky2012imagenet}}&3.4ms&2.47M\\
				{ResNet18~\cite{7780459}}&10.1ms&11.2M\\
				{MobileNet\_v2~\cite{8578572}}&13.7ms&2.2M\\
				{EfficientNet~\cite{tan2019efficientnet}}&27.4ms&39.4K\\
				{SqueezeNet1\_0~\cite{iandola2016squeezenet}}&8.8ms&735.42K\\
				{ShuffleNet\_v2\_x0.5~\cite{8578814}}&16.6ms&341.8K\\ 
				
				\bottomrule[2pt]
			\end{tabular}%
		}
		
		\label{tab:back}%
	\end{table}%
	\begin{table*}[t]
		
		\caption{Online evaluations on four benchmarks. The exhaustive comparison strongly prove the impressive real-world tracking performance. The best three performances are respectively highlighted with \textcolor[rgb]{ 1,  0,  0}{\textbf{red}}, \textcolor[rgb]{ 0,  1,  0}{\textbf{green}}, and \textcolor[rgb]{ 0,  0,  1}{\textbf{blue}} colors. Note that due to memory limitations on NVIDIA AGX Xavier, some trackers cannot run using their official code.}
		\centering
		\renewcommand\tabcolsep{8pt}
		\resizebox{1\linewidth}{!}{
			
			\begin{tabular}{lcc|ccc|ccc|cc|ccc}
				\toprule
				\multicolumn{1}{c}{\multirow{2}[0]{*}{\textbf{Tracker}}} &   \multicolumn{1}{c}{\multirow{2}[0]{*}{\textbf{Venue}}}    &    \multicolumn{1}{c|}{\multirow{2}[0]{*}{\textbf{Temporal}}}   & \multicolumn{3}{c|}{GOT-10K} & \multicolumn{3}{c|}{OTB100} & \multicolumn{2}{c|}{UAV123}  & \multicolumn{3}{c}{LaSOT} \\
				
				&  &  & \multicolumn{1}{c}{AO} & \multicolumn{1}{c}{SR$_{0.5}$} & \multicolumn{1}{c|}{SR$_{0.75}$} & \multicolumn{1}{c}{Prec.} &\multicolumn{1}{c}{Prec$_{norm}$.}& \multicolumn{1}{c|}{Succ.} & \multicolumn{1}{c}{Prec.} & \multicolumn{1}{c|}{Succ.} & \multicolumn{1}{c}{Prec.} &\multicolumn{1}{c}{Prec$_{norm}$.}& \multicolumn{1}{c}{Succ.}  \\
				\toprule
				ATOM  & \textcolor[rgb]{ .122,  .137,  .157}{CVPR 2019} & \XSolidBrush & 0.203 & 0.164 & 0.036 & 0.463 & 0.352 & 0.375 & 0.607 & 0.435 & 0.253 & 0.319 & 0.318 \\
				Dimp 18 & \textcolor[rgb]{ .122,  .137,  .157}{ICCV 2019} & \XSolidBrush & 0.221 & 0.193 & 0.039 & 0.597 & 0.458 & 0.467 & 0.700 & 0.502 & 0.322 & 0.386 & 0.377 \\
				Dimp 50 & \textcolor[rgb]{ .122,  .137,  .157}{ICCV 2019} & \XSolidBrush & 0.196 & 0.165 & 0.031 & 0.502 & 0.388 & 0.406 & 0.625 & 0.447 & 0.245 & 0.304 & 0.317 \\
				Trdimp & \textcolor[rgb]{ .122,  .137,  .157}{CVPR2021} & \Checkmark & 0.174 & 0.143 & 0.027 & 0.408 & 0.308 & 0.191 & 0.501 & 0.356 & --  &   --    & -- \\
				Prdimp18 & \textcolor[rgb]{ .122,  .137,  .157}{CVPR 2020} & \XSolidBrush & 0.169 & 0.111 & 0.026 & 0.562 & 0.416 & 0.442 & 0.689 & 0.490 & 0.321 & 0.380 & 0.378 \\
				Prdimp50 & \textcolor[rgb]{ .122,  .137,  .157}{CVPR 2020} & \XSolidBrush & 0.166 & 0.107 & 0.025 & 0.502 & 0.372 & 0.402 & 0.626 & 0.448 & 0.246 & 0.307 & 0.319 \\
				KeepTrack & \textcolor[rgb]{ .122,  .137,  .157}{ICCV 2021} & \XSolidBrush & 0.162 & 0.101 & 0.025 & 0.347 & 0.261 & 0.306 & 0.378 & 0.277 &   --    &   --    & -- \\
				KYS   & \textcolor[rgb]{ .122,  .137,  .157}{ECCV 2020} & \XSolidBrush & 0.191 & 0.160 & 0.030 & 0.475 & 0.356 & 0.386 & 0.563 & 0.398 & 0.225 & 0.287 & 0.304\\
				Tomp50 & \textcolor[rgb]{ .122,  .137,  .157}{CVPR 2022} & \XSolidBrush & 0.182 & 0.154 & 0.028 & 0.467 & 0.330 & 0.378 & 0.537 & 0.380 & 0.307 & 0.377 & 0.381 \\
				Tomp101 & \textcolor[rgb]{ .122,  .137,  .157}{CVPR 2022} & \XSolidBrush & 0.164 & 0.103 & 0.024 & 0.391 & 0.293 & 0.337 & 0.467 & 0.339 & 0.214 & 0.323 & 0.325 \\
				SiamRPN++ & \textcolor[rgb]{ .122,  .137,  .157}{CVPR2019} & \XSolidBrush & 0.239 & 0.205 & 0.047 & 0.513 & 0.343 & 0.388 & 0.519 & 0.359 & 0.277 & 0.366 & 0.363 \\
				SiamRPN++\_Mobv2 & \textcolor[rgb]{ .122,  .137,  .157}{CVPR2019} & \XSolidBrush & 0.299 & 0.287 & 0.079 & 0.688 & 0.513 & 0.518 & 0.706 & 0.483 & 0.366 & 0.436 & 0.400 \\
				SiamMask & \textcolor[rgb]{ .122,  .137,  .157}{CVPR2019} & \XSolidBrush & 0.264 & 0.242 & 0.052 & 0.613 & 0.435 & 0.461 & 0.651 & 0.447 & 0.330 & 0.419 & 0.389 \\
				SiamCAR & \textcolor[rgb]{ .122,  .137,  .157}{CVPR 2020} & \XSolidBrush & 0.243 & 0.206 & 0.047 & 0.528 & 0.359 & 0.404 & 0.522 & 0.358 & 0.291 & 0.358 & 0.352 \\
				SiamBAN & \textcolor[rgb]{ .122,  .137,  .157}{CVPR 2020} & \XSolidBrush & 0.249 & 0.212 & 0.058 & 0.508 & 0.346 & 0.397 & 0.527 & 0.365 & 0.300 & 0.393 & 0.377 \\
				SiamAPN & \textcolor[rgb]{ .122,  .137,  .157}{ICRA2021} & \XSolidBrush & 0.293 & 0.270 & 0.063 & \textcolor[rgb]{ 0,  0,  1}{\textbf{0.701}} & \textcolor[rgb]{ 0,  1,  0}{\textbf{0.540}} & \textcolor[rgb]{ 0,  1,  0}{\textbf{0.529}} & \textcolor[rgb]{ 0,  0,  1}{\textbf{0.729}} & 0.505 & \textcolor[rgb]{ 0,  0,  1}{\textbf{0.383}} & \textcolor[rgb]{ 0,  1,  0}{\textbf{0.461}} & 0.398 \\
				SiamAPN++ & \textcolor[rgb]{ .122,  .137,  .157}{IROS2021} & \XSolidBrush & 0.304 & 0.286 & 0.064 & 0.683 & 0.532 & \textcolor[rgb]{ 0,  0,  1}{\textbf{0.522}} & 0.717 & \textcolor[rgb]{ 0,  1,  0}{\textbf{0.518}} & \textcolor[rgb]{ 0,  1,  0}{\textbf{0.386}} & 0.449 & 0.397 \\
				HiFT  & \textcolor[rgb]{ .122,  .137,  .157}{ICCV2021} & \XSolidBrush & \textcolor[rgb]{ 0,  1,  0}{\textbf{0.328}} & \textcolor[rgb]{ 0,  1,  0}{\textbf{0.331}} & 0.071 & 0.655 & 0.506 & 0.494 & 0.717 & 0.497 & 0.364 & 0.448 & \textcolor[rgb]{ 0,  0,  1}{\textbf{0.418}} \\
				SiamGAT & \textcolor[rgb]{ .122,  .137,  .157}{CVPR2021} & \XSolidBrush & 0.314 & 0.310 & \textcolor[rgb]{ 0,  0,  1}{\textbf{0.086}} & 0.688 & 0.499 & 0.514 & 0.695 & 0.481 & 0.318 & 0.423 & 0.405 \\
				TransT & \textcolor[rgb]{ .122,  .137,  .157}{CVPR2021} & \XSolidBrush & \textcolor[rgb]{ 0,  0,  1}{\textbf{0.322}} & \textcolor[rgb]{ 0,  0,  1}{\textbf{0.314}} & \textcolor[rgb]{ 0,  1,  0}{\textbf{0.103}} & 0.597 & 0.414 & 0.450 & 0.658 & 0.453 & 0.339 & 0.424 & \textcolor[rgb]{ 0,  1,  0}{\textbf{0.434}} \\
				TCTrack & \textcolor[rgb]{ .122,  .137,  .157}{Ours} & \Checkmark & 0.312 & 0.291 & 0.069 & \textcolor[rgb]{ 0,  1,  0}{\textbf{0.704}} & \textcolor[rgb]{ 0,  0,  1}{\textbf{0.537}} & 0.518 & \textcolor[rgb]{ 0,  1,  0}{\textbf{0.730}} & \textcolor[rgb]{ 0,  0,  1}{\textbf{0.510}} & 0.381 & \textcolor[rgb]{ 0,  0,  1}{\textbf{0.451}} & 0.409 \\
				\midrule
				\textbf{TCTrack++} & \textbf{Ours} & \textbf{\Checkmark} & \textcolor[rgb]{ 1,  0,  0}{\textbf{0.375}} & \textcolor[rgb]{ 1,  0,  0}{\textbf{0.394}} & \textcolor[rgb]{ 1,  0,  0}{\textbf{0.117}} & \textcolor[rgb]{ 1,  0,  0}{\textbf{0.720}} & \textcolor[rgb]{ 1,  0,  0}{\textbf{0.550}} & \textcolor[rgb]{ 1,  0,  0}{\textbf{0.543}} & \textcolor[rgb]{ 1,  0,  0}{\textbf{0.731}} & \textcolor[rgb]{ 1,  0,  0}{\textbf{0.519}} & \textcolor[rgb]{ 1,  0,  0}{\textbf{0.414}} & \textcolor[rgb]{ 1,  0,  0}{\textbf{0.484}} & \textcolor[rgb]{ 1,  0,  0}{\textbf{0.435}} \\
				
				\bottomrule
			\end{tabular}%
			
		}
		
		\label{tab:gotlasot}%
	\end{table*}%

	\section{Experiments}\label{sec:Evaluation}
	
	In this section, our tracker is evaluated on eight public authoritative benchmarks, \textit{i.e.}, OTB100~\cite{otb}, GOT-10K~\cite{huang2019got}, LaSOT~\cite{fan2019lasot}, UAV123~\cite{Mueller2016ECCV}, UAVTrack112\_L~\cite{siamapnj}, UAV123@10fps~\cite{Mueller2016ECCV}, VisDrone2018-test~\cite{wen2018visdrone}, and DTB70~\cite{li2017visual} and tested on real-world tracking conditions. Besides, more than 50 existing top trackers are included for a thorough comparison, where their results are obtained by running the official codes with their corresponding hyper-parameters. Overall, our evaluation are divided into two groups: \textbf{(i)} online evaluation to comprehensively evaluate the real-world tracking performance of all trackers; \textbf{(ii)} offline evaluation to demonstrate the tracking performance of trackers without latency. For a clearer comparison of offline evaluation, we also divide those trackers into two groups, \textbf{(i)} light-weight trackers~\cite{kcf,bertinetto2016staple,bacf,eco,DSST,strcf,kcc,Huang2019ICCV,Li_2020_CVPR,dasiamrpn,siamrpn++,xu2020siamfc++,siamapn,hift,lighttrack,cao2022tctrack} and \textbf{(ii)} deep trackers~\cite{mixformer,tomp,ulast,siamcar,keeptrack,trdimp,transt,stmtrack,siamban,prdimp,kys,siamattn,siamrcnn,atom,dimp,siamrpn++,siamdw,tadt,siamrpn,traca,strucsiam,siamfctri,cfnet,siamfc,DSST,struck,lukezic2020d3s,sesiamfc,ocean,updatenet,siamgat,siammask,gao2022aiatrack}. For distinguishing the tracker with different backbone, we define SiamRPN++\_A (SiamRPN++ equipped with AlexNet) and SiamRPN++\_Mobv2 (SiamRPN++ equipped with MobileNet\_v2). Codes and models are available at \url{https://github.com/vision4robotics/TCTrack}.

	\begin{figure*}[t]
		\centering	
		\subfloat[Aspect Ration Change on LaSOT]
		{
			\includegraphics[width=0.32\textwidth]{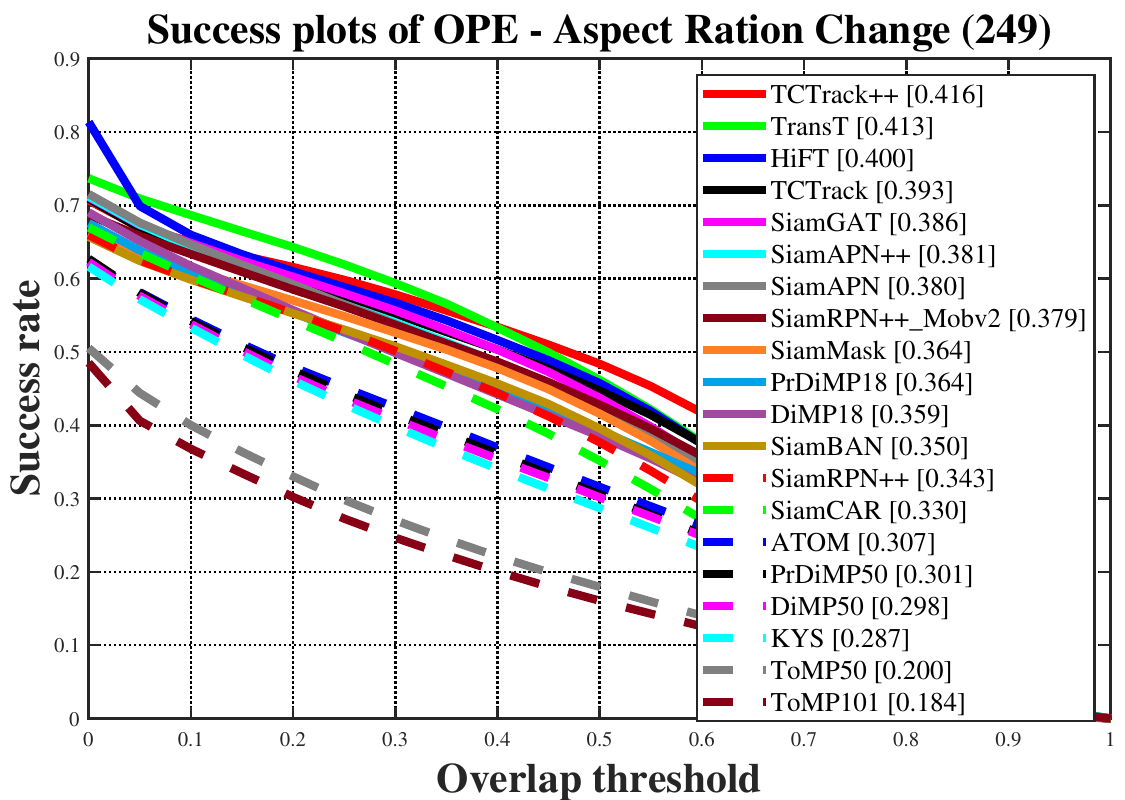}
		}
		\subfloat[Camera Motion on LaSOT]
		{
			\includegraphics[width=0.32\textwidth]{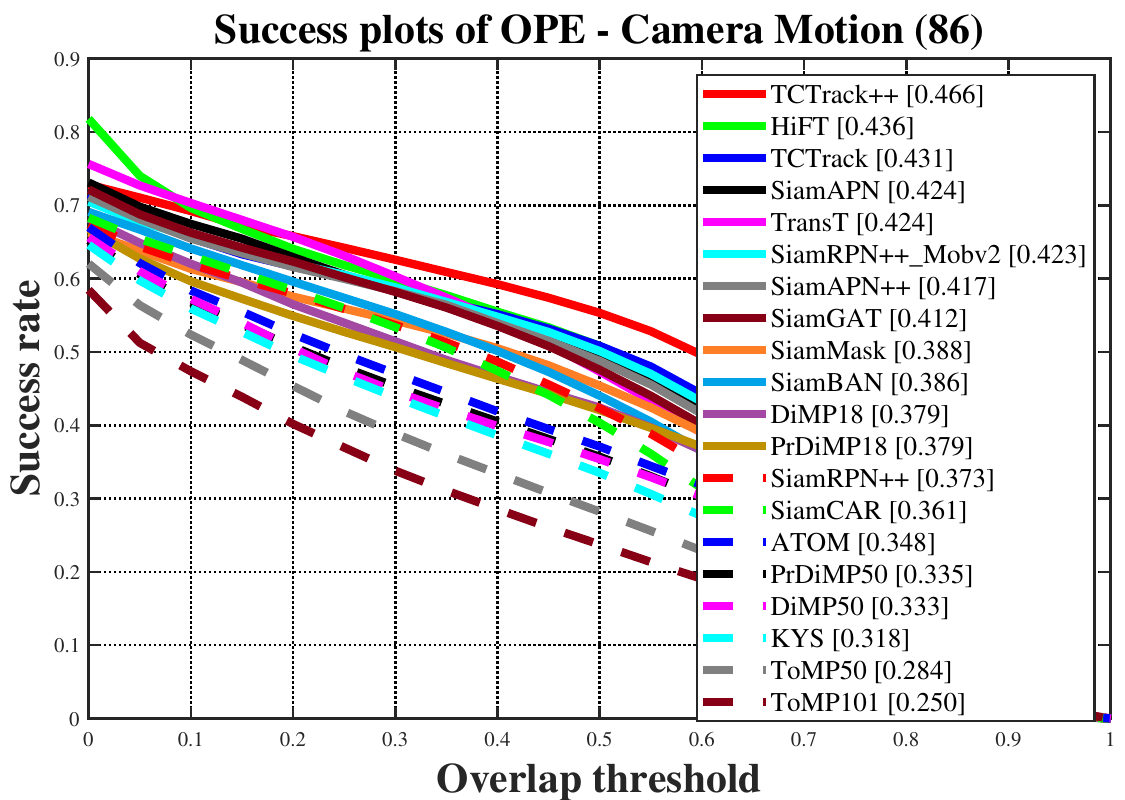}
		}
		\subfloat[Fast Motion on LaSOT]
		{
			\includegraphics[width=0.32\textwidth]{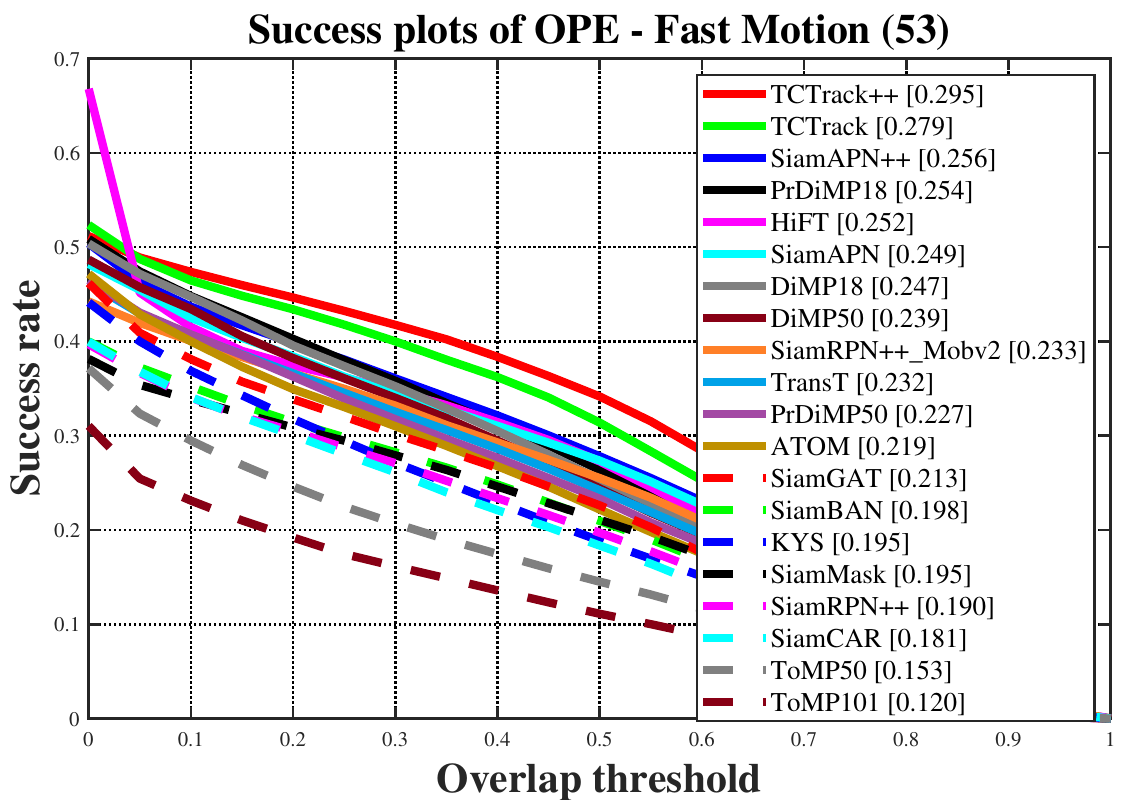}
		}
		\\
		
		\subfloat[Full Occlusion on LaSOT]
		{
			\includegraphics[width=0.32\textwidth]{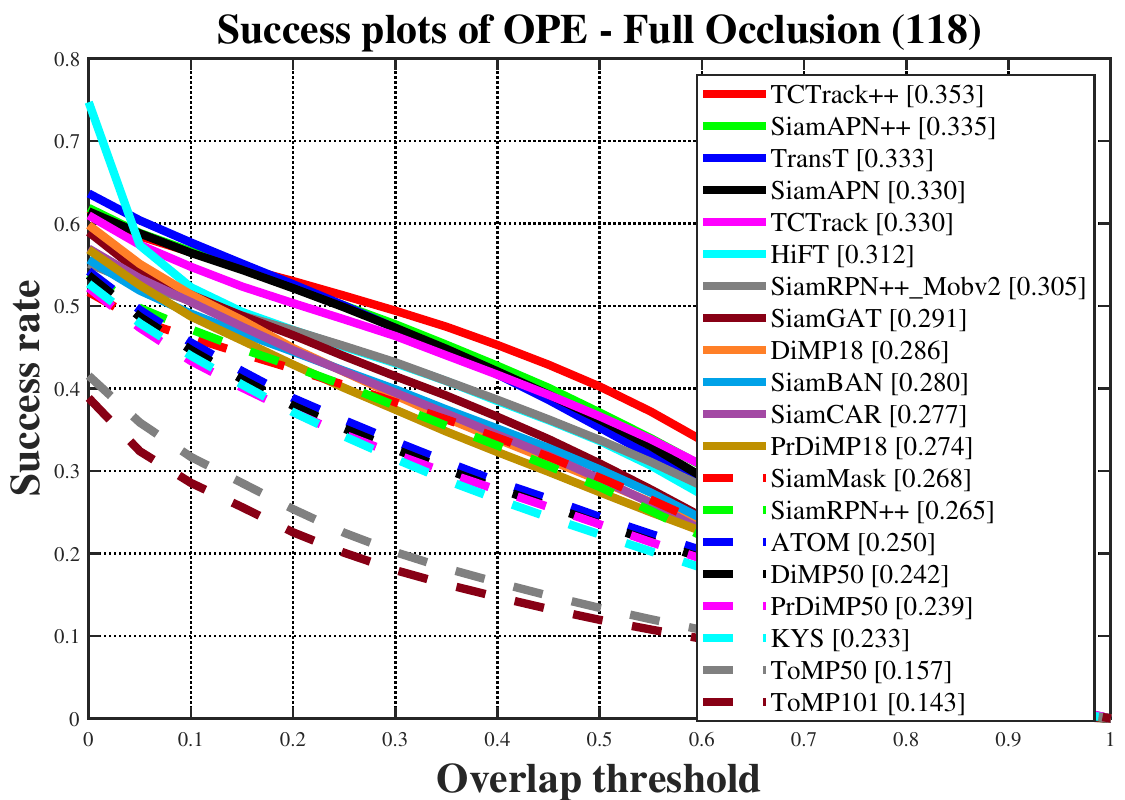}
		}
		\subfloat[Illumination Variation on LaSOT]
		{
			\includegraphics[width=0.32\textwidth]{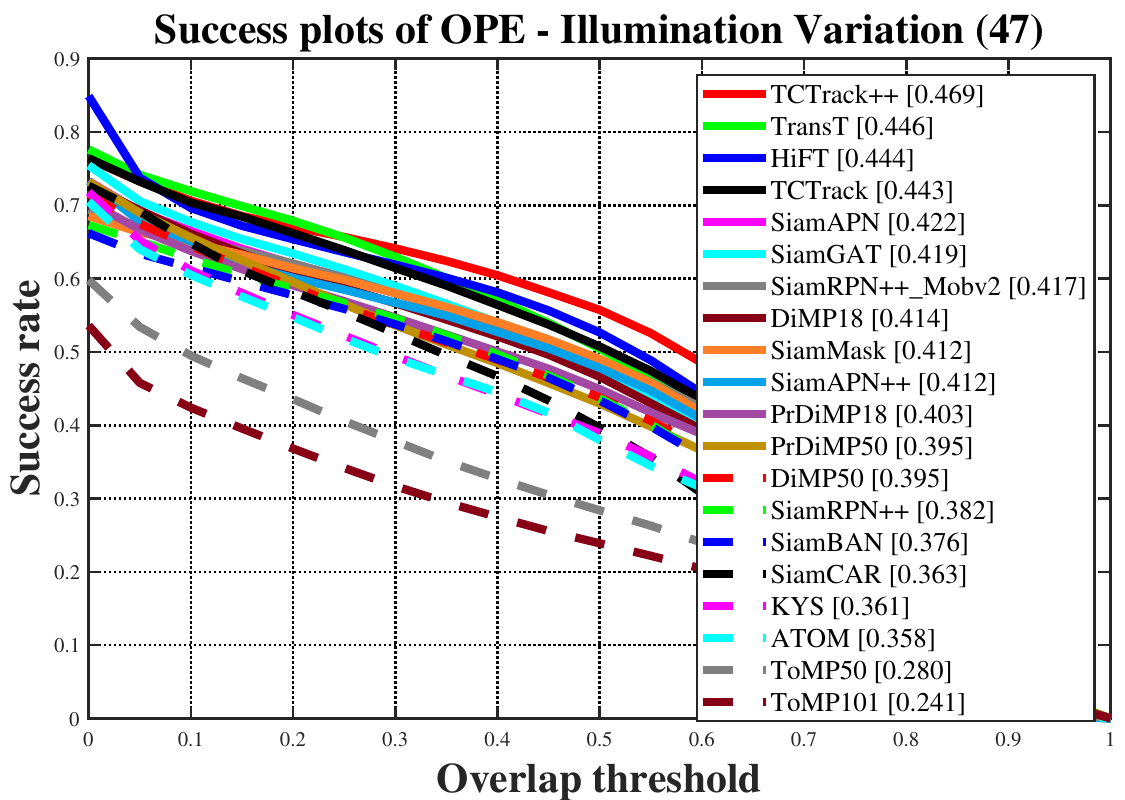}
		}
		\subfloat[Low Resolution on LaSOT]
		{
			\includegraphics[width=0.32\textwidth]{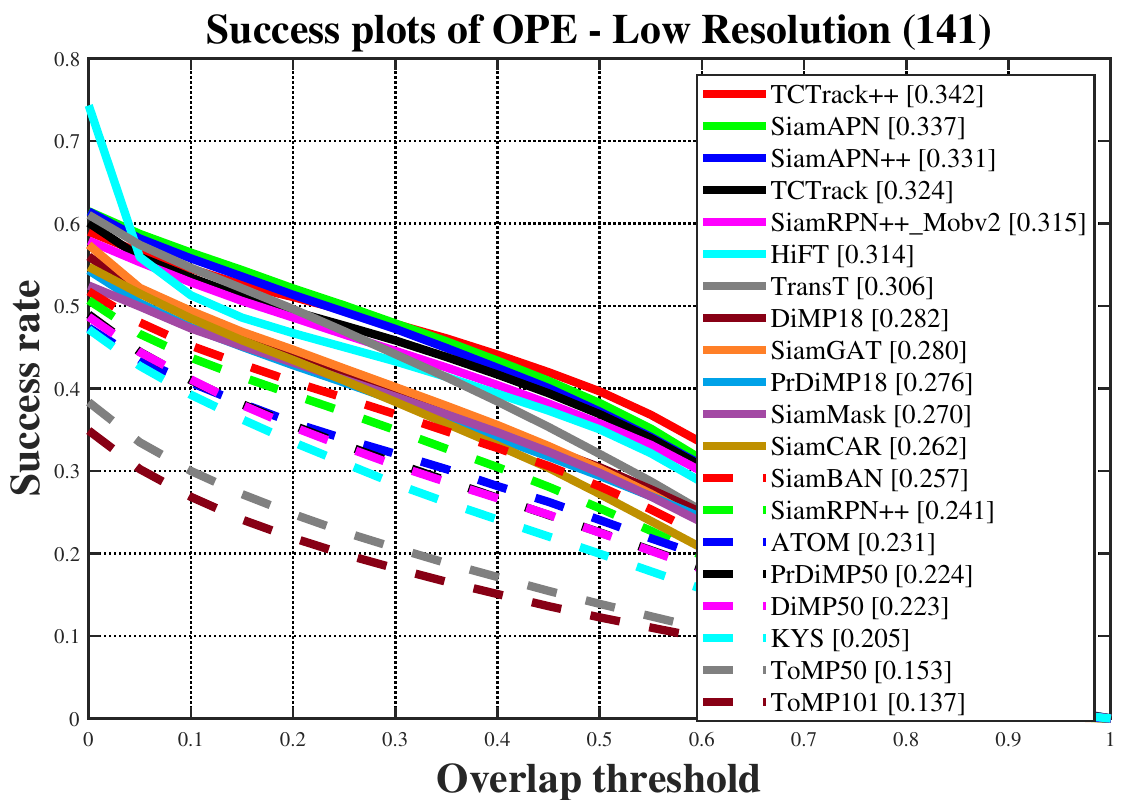}
		}
		\\
		
		\subfloat[Motion Blur on LaSOT]
		{
			\includegraphics[width=0.32\textwidth]{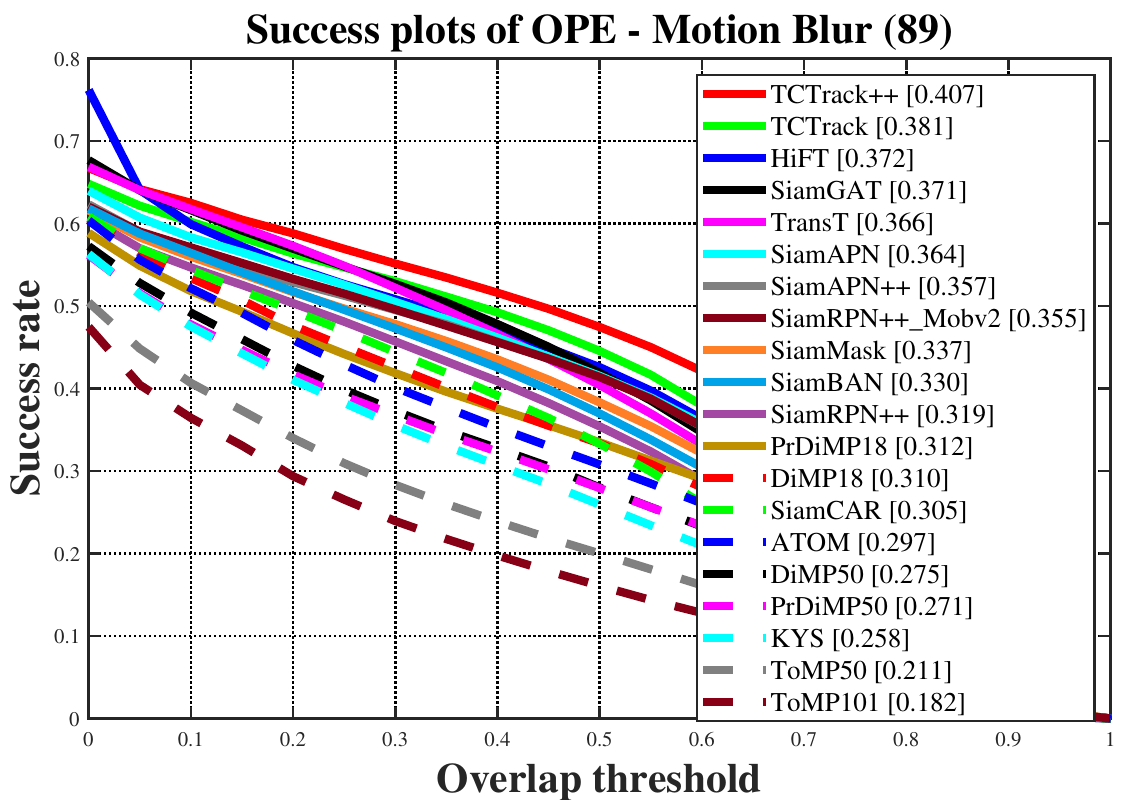}
		}
		\subfloat[Out-of-View on LaSOT]
		{
			\includegraphics[width=0.32\textwidth]{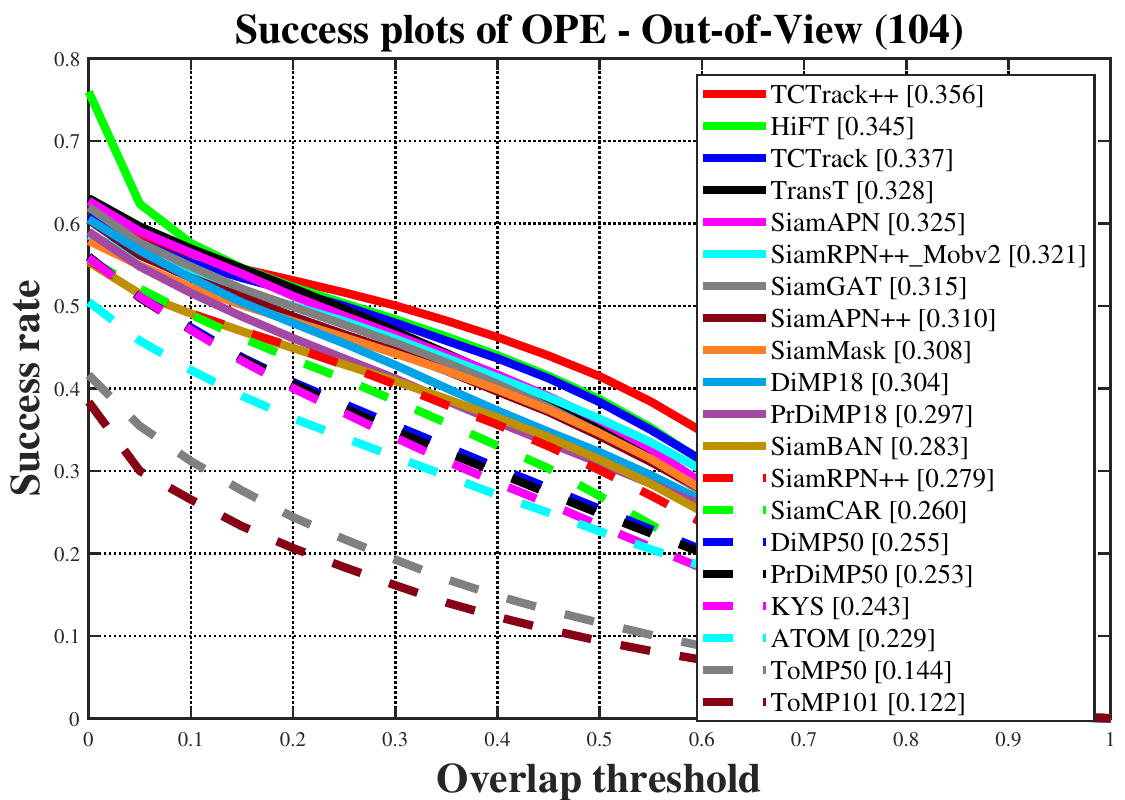}
		}
		\subfloat[Viewpoint Change on LaSOT]
		{
			\includegraphics[width=0.32\textwidth]{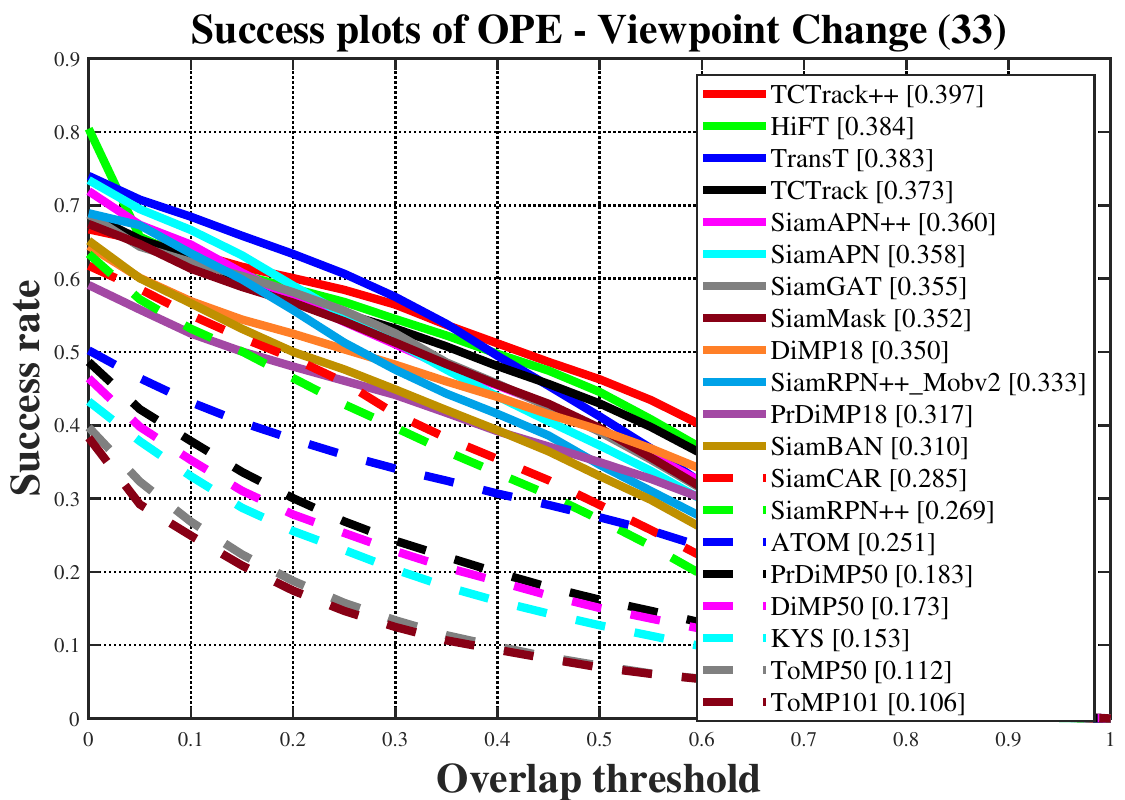}
		}
		\\

		\caption{Attribute-based evaluation of all trackers on LaSOT. Our temporal tracker can maintain promising performance under (a) ARC, (b) CM, (c) FM, (d) FOC, (e) IV, (f) LR, (g) MB, (h) OV, and (i) VC scenes. }\label{fig:att}
		
	\end{figure*}
	
	\subsection{Implementation Details}
	
	To meet the requirement of real-world tracking, we conduct the comparison of the inference time of popular lightweight backbones on the resource-limited platform. AlexNet has the lowest latency as shown in TABLE~\ref{tab:back}, while the recent developments in mobile networks~\cite{iandola2016squeezenet,8578572,8578814} suffer from high memory access cost (MAC). For clarification, we denote the inference time of the tracker as $\tau$. Ideally, when the real world time is $t$ the tracker will provide $t$-th frame prediction results without latency. However, in real-world tracking, the tracker can get results of $t$-th frame results when the time goes to $t+\tau$. Therefore, higher inference time will enlarge the mismatch between real-world information and tracker, significantly blocking the real-world tracking performance. Thus, we choose the AlexNet as the backbone of our tracker.


	For initialization, we use the ImageNet pre-trained model for AlexNet and use the same initialization for ATT-TAdaConv as in~\cite{russakovsky2015imagenet}. The AT-Trans in our TCTrack++ is randomly initialized. We train our tracker with the videos whose lengths are transformed from 1 to 4 via curriculum learning strategy. Training images are extracted from COCO~\cite{lin2014microsoft}, VID~\cite{russakovsky2015imagenet}, LaSOT~\cite{fan2019lasot}, and GOT-10K~\cite{huang2019got}. Our tracker is trained on two NVIDIA TITAN RTX GPUs with 100 epochs. For the first 10 epochs, the parameters of the backbone are frozen while for the last 90 epochs all parameters will be updated. The training process employs a learning rate decreasing from $0.005$ to $0.0005$ in log space. 
	We adopt SGD as the optimizer with a momentum of $0.9$, where the mini-batch size is $124$ pairs.
	The input sizes of the template and the search area are $127^2$ and $287^2$ respectively. The proposed ATT-TAdaConv is used in the replacement of the last two convolutional layers. Note that because of our curriculum learning strategy, the video length is set to 2 at the first 33 epochs. During the 33 to 50 epochs, the video length is varied to 3. After that, the video length will increase to 4.
	
	
	\subsection{Evaluation Criteria}
	
	In this paper, we mainly adopt the one-pass evaluation (OPE) criteria to evaluate the performance, \textit{i.e.}, precision and success rate. The former is defined by the center location error (CLE) between the ground truth bounding box and the predicted box while the latter is calculated by Intersection over Union (IoU) score. Note that we use the area-under-the-cure (AUC) as the metric to rank the trackers. Additionally, average overleap (AO) and success rate (SR) are adopted in GOT-10K evaluations. Similarly, AO is the average IoU score between all predicted boxes and ground truth while SR$_{i}$ represents the rate of frames whose IoU score is higher than $i$.

	\begin{figure*}[t]
		\centering
		\includegraphics[width=0.99\linewidth]{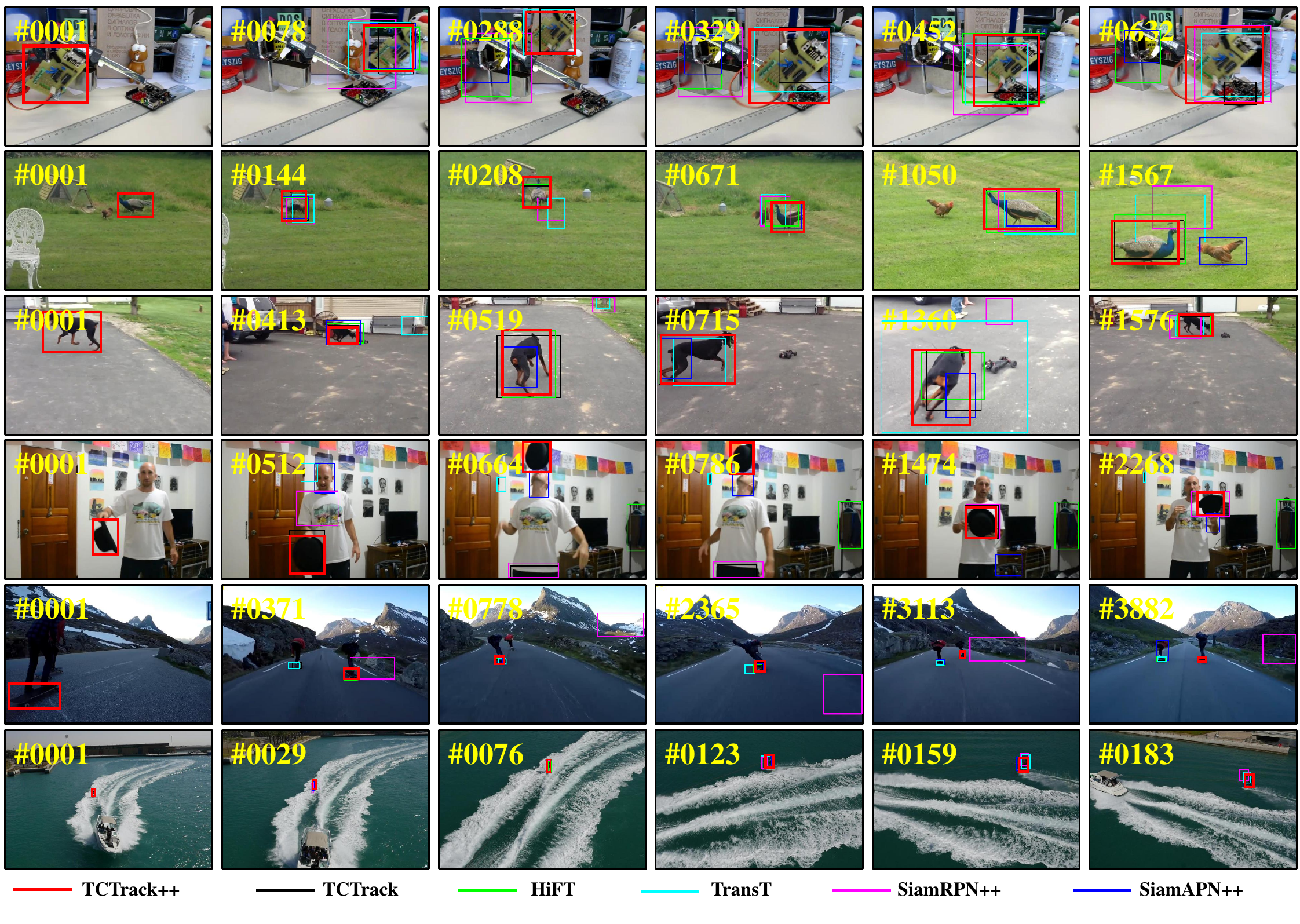}
		\vspace{-10pt}
		\caption{Qualitative comparisons among TCTrack++ and other 5 trackers, \textit{i.e.}, TransT~\cite{transt}, HiFT~\cite{hift}, TCTrack~\cite{cao2022tctrack}, SiamRPN++~\cite{siamrpn++}, and  SiamAPN++~\cite{siamapn++}. From the up to bottom, the sequences are from OTB100~\cite{otb} (\textit{Board}), LaSOT~\cite{fan2019lasot} (\textit{bird-3}, \textit{dog-7}, \textit{hat-18}, and \textit{skateboard-8}), and UAV123~\cite{Mueller2016ECCV} (\textit{wakeboard7}). It clearly shows that our tracker achieves superior performance in various challenging scenarios. Best viewed on the screen with high-resolution.}
		
		\label{fig:quali} 
	\end{figure*}
			\begin{table*}[t]
		\centering
		\caption{Comparisons to trackers with light-weight trackers on three benchmarks. Note that GPU* represents the GPU on the embedded system NVIDIA Jetson AGX Xavier. Our tracker achieves superior performance compared with other light-weight trackers. The best three
			performances are respectively highlighted with \textcolor[rgb]{ 1,  0,  0}{\textbf{red}}, \textcolor[rgb]{ 0,  1,  0}{\textbf{green}}, and \textcolor[rgb]{ 0,  0,  1}{\textbf{blue}} colors.}
		\renewcommand\tabcolsep{4.5pt}
		\begin{tabular}{lcccc| ccc|ccc|ccc}
			\toprule
			
			\multicolumn{1}{c}{\multirow{2}[0]{*}{\textbf{Tracker}}} & \multicolumn{1}{c}{\multirow{2}[0]{*}{\textbf{Venue}}} &\multicolumn{1}{c}{\multirow{2}[0]{*}{\textbf{Temporal}}}& \multicolumn{1}{c}{\multirow{2}[0]{*}{\textbf{Platform}}} & \multicolumn{1}{c|}{\multirow{2}[0]{*}{\textbf{FPS}}} & \multicolumn{3}{c|}{\textbf{UAV123@10fps}} & \multicolumn{3}{c|}{\textbf{UAVTrack112}} & \multicolumn{3}{c}{\textbf{VisDrone2018}} \\

			&     &  &  &  & \multicolumn{1}{c}{Prec.} & \multicolumn{1}{c}{Prec$_{norm}$.} & \multicolumn{1}{c|}{Succ.} & \multicolumn{1}{c}{Prec.} & \multicolumn{1}{c}{Prec$_{norm}$.} & \multicolumn{1}{c|}{Succ.} & \multicolumn{1}{c}{Prec.} & \multicolumn{1}{c}{Prec$_{norm}$.} & \multicolumn{1}{c}{Succ.} \\
			\midrule
			
			DCF   & TPAMI2015 &\XSolidBrush& CPU   & \textcolor[rgb]{ 1,  0,  0}{\textbf{811.45}} & 0.408 & 0.348 & 0.266 & 0.421 & 0.350 & 0.264 & 0.669 & 0.498 & 0.408 \\
			KCF   & TPAMI2015 &\XSolidBrush& CPU   & \textcolor[rgb]{ 0,  1,  0}{\textbf{561.09}} & 0.406 & 0.346 & 0.265 & 0.421 & 0.352 & 0.266 & 0.685 & 0.512 & 0.413 \\
			Staple & CVPR2016 &\XSolidBrush& CPU   & 56.05 & 0.456 & 0.398 & 0.342 & 0.624 & 0.512 & 0.415 & 0.783 & 0.629 & 0.566 \\
			
			BACF  & ICCV2017 &\XSolidBrush& CPU   & 52.52 & 0.572 & 0.484 & 0.413 & 0.611 & 0.498 & 0.408 & 0.774 & \textcolor[rgb]{ 0,  0,  1}{\textbf{0.632}} & 0.567 \\
			
			ECO-HC & CVPR2017 &\XSolidBrush& CPU   & 66.58 & \textcolor[rgb]{ 0,  0,  1}{\textbf{0.634}} & \textcolor[rgb]{ 0,  0,  1}{\textbf{0.538}} & \textcolor[rgb]{ 0,  0,  1}{\textbf{0.462}} &  \textcolor[rgb]{ 0,  1,  0}{\textbf{0.686}} & \textcolor[rgb]{ 0,  1,  0}{\textbf{0.545}} & \textcolor[rgb]{ 1,  0,  0}{\textbf{0.474}} & \textcolor[rgb]{ 0,  1,  0}{\textbf{0.797}} & 0.620  & \textcolor[rgb]{ 0,  1,  0}{\textbf{0.578}} \\
			fDSST & TPAMI2017 &\XSolidBrush& CPU   & \textcolor[rgb]{ 0,  0,  1}{\textbf{153.66}} & 0.516 & 0.439 & 0.379 & 0.568 & 0.456 & 0.391 & 0.698 & 0.559 & 0.51 \\

			
			Staple\_CA & CVPR2017 &\XSolidBrush& CPU   & 62.05 & 0.587 & 0.500   & 0.420  & 0.634 & 0.522 & 0.426 & 0.782 & 0.615 & 0.550 \\
			STRCF & CVPR2018 &\Checkmark& CPU   & 26.93 & 0.627 & 0.536 & 0.457 & 0.645 & 0.518 & 0.430 & 0.778 & \textcolor[rgb]{ 0,  1,  0}{\textbf{0.636}} & 0.567 \\
			KCC   & AAAI2018 &\XSolidBrush& CPU   & 40.81 & 0.531 & 0.461 & 0.374 & 0.572 & 0.474 & 0.379 & 0.781 & 0.610  & 0.530 \\
			ARCF & ICCV2019 &\Checkmark& CPU   & 24.05 & \textcolor[rgb]{ 0,  1,  0}{\textbf{0.666}} & \textcolor[rgb]{ 0,  1,  0}{\textbf{0.559}} & \textcolor[rgb]{ 0,  1,  0}{\textbf{0.473}} & \textcolor[rgb]{ 0,  0,  1}{\textbf{0.672}} & \textcolor[rgb]{ 0,  1,  0}{\textbf{0.545}} & \textcolor[rgb]{ 0,  0,  1}{\textbf{0.456}} & \textcolor[rgb]{ 0,  0,  1}{\textbf{0.794}} & \textcolor[rgb]{ 1,  0,  0}{\textbf{0.642}} & \textcolor[rgb]{ 1,  0,  0}{\textbf{0.587}} \\
			Autotrack & CVPR2020 &\Checkmark& CPU   & 47.61 & \textcolor[rgb]{ 1,  0,  0}{\textbf{0.671}} & \textcolor[rgb]{ 1,  0,  0}{\textbf{0.562}} & \textcolor[rgb]{ 1,  0,  0}{\textbf{0.477}} & \textcolor[rgb]{ 1,  0,  0}{\textbf{0.694}} & \textcolor[rgb]{ 1,  0,  0}{\textbf{0.556}} & \textcolor[rgb]{ 0,  1,  0}{\textbf{0.465}} & 0.777 & 0.614 & 0.552 \\
			
			\midrule
			DaSiamRPN & ECCV2018 &\XSolidBrush& GPU*  & \textcolor[rgb]{ 1,  0,  0}{\textbf{44}} & 0.692 & 0.542 & 0.482 & 0.702 & 0.544 & 0.48  & 0.72  & 0.579 & 0.526 \\
			SiamRPN++\_A & CVPR2019 &\XSolidBrush& GPU*  & \textcolor[rgb]{ 0,  1,  0}{\textbf{41}} & 0.735 & 0.631 & 0.55  & \textcolor[rgb]{ 0,  0,  1}{\textbf{0.793}} & 0.677 & 0.592 & \textcolor[rgb]{ 0,  1,  0}{\textbf{0.786}} & \textcolor[rgb]{ 0,  0,  1}{\textbf{0.639}} & \textcolor[rgb]{ 0,  0,  1}{\textbf{0.568}} \\
			SiamFC++\_A & AAAI2020 &\XSolidBrush& GPU*  & 34    & 0.745 & 0.657 & 0.576 & 0.77  & \textcolor[rgb]{ 0,  0,  1}{\textbf{0.684}} & \textcolor[rgb]{ 0,  0,  1}{\textbf{0.608}} & \textcolor[rgb]{ 0,  0,  1}{\textbf{0.773}} & \textcolor[rgb]{ 0,  1,  0}{\textbf{0.655}} & \textcolor[rgb]{ 0,  1,  0}{\textbf{0.577}} \\
			SiamAPN++ & IROS2021 &\XSolidBrush& GPU*  & \textcolor[rgb]{ 0,  0,  1}{\textbf{34.9}} & 0.764 & 0.669 & 0.58  & 0.769 & 0.665 & 0.585 & 0.732 & 0.604 & 0.53 \\
			HiFT  & ICCV2021 &\XSolidBrush& GPU*  & 31.2  & 0.749 & 0.655 & 0.569 & 0.742 & 0.644 & 0.57  & 0.712 & 0.593 & 0.521 \\
			LightTrack & CVPR2021 &\XSolidBrush& GPU*  & 19    & \textcolor[rgb]{ 0,  1,  0}{\textbf{0.776}} & \textcolor[rgb]{ 0,  0,  1}{\textbf{0.673}} & \textcolor[rgb]{ 1,  0,  0}{\textbf{0.599}} & \textcolor[rgb]{ 0,  1,  0}{\textbf{0.811}} & \textcolor[rgb]{ 0,  1,  0}{\textbf{0.692}} & \textcolor[rgb]{ 0,  1,  0}{\textbf{0.638}} & 0.742 & 0.618 & 0.564 \\
			TCTrack & CVPR2022 &\Checkmark& GPU*  & 27.4    & \textcolor[rgb]{ 0,  0,  1}{\textbf{0.774}} & \textcolor[rgb]{ 0,  1,  0}{\textbf{0.675}} & \textcolor[rgb]{ 0,  0,  1}{\textbf{0.587}} & 0.766 & 0.659 & 0.594 & 0.765 & 0.633 & \textcolor[rgb]{ 0,  0,  1}{\textbf{0.568}} \\
			\textbf{TCTrack++} & Ours  &\Checkmark& GPU*  & 27.1    & \textcolor[rgb]{ 1,  0,  0}{\textbf{0.780}} & \textcolor[rgb]{ 1,  0,  0}{\textbf{0.681}} & \textcolor[rgb]{ 1,  0,  0}{\textbf{0.599}} & \textcolor[rgb]{ 1,  0,  0}{\textbf{0.815}} & \textcolor[rgb]{ 1,  0,  0}{\textbf{0.707}} & \textcolor[rgb]{ 1,  0,  0}{\textbf{0.640}} & \textcolor[rgb]{ 1,  0,  0}{\textbf{0.799}} & \textcolor[rgb]{ 1,  0,  0}{\textbf{0.658}} & \textcolor[rgb]{ 1,  0,  0}{\textbf{0.594}} \\
			\bottomrule
		\end{tabular}
		\label{tab:addlabel}%
	\end{table*}%
	\subsection{Online Evaluation}\label{sec:online eva}
	
	As discussed before, the tracking performance in real-world scenes is influenced by the latency caused by the algorithm. Therefore, to more comprehensively and realistically evaluate the tracking performance in real-world scenes, we conduct the latency-aware evaluation by considering the inference time during tracking on the resource-limited platform, \textit{i.e.}, NVIDIA Jetson AGX Xavier. 
	Please note that our online evaluations are conducted on one hardware to make sure that we can control all other factors that can affect tracking results. Besides, all trackers are adopted their official code.
	\subsubsection{Evaluation Settings}
	Inspired by~\cite{li2020towards}, latency-aware online evaluations are proposed~\cite{li2021predictive} by considering the effect of inference time. Different from standard offline evaluations, it couples the speed and the performance together. The key point is that the tracker must provide the latest results in every time node. For instance, when the tracker receives the $t$-th frame, the tracker must provide the tracking result immediately. However, due to the latency of the algorithm ($\tau$), the latest tracking result is calculated by $(t-\tau)$-th frame. Therefore, no matter how fast the tracker it is, it will be a small mismatch between the tracker and the real world. 
	
	\subsubsection{Results on OTB100} 
	
	OTB100~\cite{otb} has become one of the most common benchmarks for evaluating tracking performance, containing 100 challenging scenarios. Besides, we adopt normalized precision plots, precision plots and success plots in one-pass evaluation (OPE) to evaluate tracking performance. As shown in TABLE.~\ref{tab:gotlasot}, our tracker TCTrack++ ranks first in precision (0.720), normalized precision (0.550), and success rate (0.543). Compared with the best deep trackers, our tracker makes a 4.7\%, 7.2\%, and 4.8\% improvement in three metrics. Relying on the efficient and effective modules, our tracker achieves impressive performance in online evaluations.

	\subsubsection{Results on LaSOT} 
	
	We also adopt LaSOT~\cite{fan2019lasot} as one of our evaluation benchmarks. It collects more than 3.5 million frames and consists of 1400 sequences. Each video has more than 2500 frames on average. Also, it builds several attributes, \textit{e.g.}, aspect ration change (ARC), camera motion (CM), fast motion (FM), full occlusion (FOC), illumination variation (IV), low resolution (LR), motion blur (MB), out-of-view (OV), viewpoint change (VC) and so on. 
	
	The overall comparison results are released in TABLE.~\ref{tab:gotlasot}. Benefiting from our efficient temporal framework, our tracker can exploit the temporal contexts while introducing as low latency as possible. Thus, compared with the best deep tracker, our TCTrack++ achieves considerable improvement in precision (\textbf{13.1\%}) and normalized precision (\textbf{11.0\%}), respectively. Furthermore, our tracker surpasses other light-weight trackers and our previous version TCTrack by \textbf{7.3\%} and \textbf{8.7\%}.

	We also conduct the attribute-based evaluations for verifying the robustness of our tracker in different challenging scenes, which are shown in Fig.~\ref{fig:att}. Our tracker maintains the 1-st rank in 9 attributes including ARC, CM, FM, FOC, IV, LR, MB, OV, and VC. Especially, TCTrack++ exceeds the second-best tracker with an improvement of \textbf{6.9\%}.
	
	In summary, the exhaustive experiments on such a large-scale benchmark strongly validate the superior real-world tracking performance of our tracker.

	\subsubsection{Results on UAV123} 
	
	By adopting the unmanned aerial vehicle (UAV) to record videos, UAV123~\cite{Mueller2016ECCV} collects 123 challenging sequences with more than 112K frames. The maximum video length is 3085. The fast motion of the UAV introduces severe motion blur and camera motion, which have a significant influence on visual trackers.
	
	TABLE~\ref{tab:gotlasot} illustrates the comparison of our TCTrack++ against other deep trackers including ToMP~\cite{tomp}, TransT~\cite{transt}, KeepTrack~\cite{keeptrack}, and TrDiMP~\cite{trdimp}. Despite excellent tracking performance in offline evaluation, their high latency structure will affect the tracking performance in real-world tracking conditions, especially in fast-motion aerial scenarios. Thanks to our efficient temporal framework, our TCTrack++ can maintain an impressive trade-off between performance and speed. Therefore, our tracker achieves the 1-st rank in both precision (0.731) and success (0.519).

	\subsubsection{Results on GOT-10K} 
	GOT-10K~\cite{huang2019got} is a large-scale visual tracking benchmark in the wild, which contains more than 10K video segments of real-world moving objects and over 1.5 million manually labeled bounding boxes. For a fair comparison, the results in TABLE~\ref{tab:gotlasot} are collected from their official website.

	TABLE~\ref{tab:gotlasot} shows that even compared with the most existing SOTA deep tracker ToMP~\cite{tomp} (CVPR2022), our tracker achieves significant improvements in real-world tracking conditions. Besides, TCTrack++ outperforms the second-best trackers by \textbf{14.3\%} in AO, \textbf{19.0\%} in SR$_{0.5}$, and \textbf{13.6\%} in SR$_{0.75}$. 
	
	\subsubsection{Qualitative Results}
	
	For demonstrating the real-world tracking performance intuitively, the tracking results are visualized as shown in Fig.~\ref{fig:quali}. The qualitative results on several challenging scenes show that compared with the other 5 trackers, our TCTrack++ maintains promising robustness and accuracy in FM, SV, OC, and so on.

	\subsection{Offline Evaluation}

	Additionally, to exhaustively evaluate the tracking performance, we construct the offline evaluations on five benchmarks. Note that we follow the standard pipeline for offline evaluations. For clarification, we will evaluate the light-weight trackers and deep trackers respectively.

	\subsubsection{Comparison with Light-Weight Trackers}~\label{over}
	In this subsection, TCTrack is compared with existing efficient trackers on several benchmarks except for some methods without code including 1) real-time speed on CPU; 2) real-time speed on embedded GPU (NVIDIA Jetson AGX Xavier).
	


	\noindent\textbf{UAV123@10fps.} Downsampling from the 30FPS version, UAV123@10fps~\cite{Mueller2016ECCV} merely has 10 frames per second. Therefore, the motion among consecutive frames is more severe than the sequences of 30FPS. From the comparison with other light-weight trackers shown in TABLE~\ref{tab:addlabel}, we can find that our tracker achieves real-time speed and the best tracking performance. Besides, compared with the baseline TCTrack, our tracker gain a 2\% improvement in success rate.

	\noindent\textbf{UAVTrack112.} 
	UAVTrack112~\cite{siamapnj} contains 112 challenging aerial scenarios with more than 100K frames in total covering UAVTrack112\_L. Especially, there are some low illumination sequences collected in the dark time, which is very challenging for visual trackers. The superior evaluation results of our TCTrack++ prove the impressive accuracy and robustness compared with other lightweight trackers.

	\noindent\textbf{VisDrone2018-test.} 
	VisDrone2018-test~\cite{wen2018visdrone} includes 35 challenging scenes which involve urban, country, pedestrian, vehicles, sparse \& crowded people, \textit{etc}. From the comparison, we can clearly find that compared with our baseline TCTrack~\cite{cao2022tctrack} and the most recent tracker LightTrack~\cite{lighttrack}, our TCTrack++ maintains the promising robustness and achieves 1-st rank on three benchmarks.

	\subsubsection{Comparison with Deep SOTA Trackers}\label{Sec:deep}
	
	To validate the superiority of our tracker, we evaluate our tracker with the most recent deep SOTA trackers.
	
\begin{figure}[t]
	\centering	
	\vspace{-10pt}
	\includegraphics[width=0.5\textwidth]{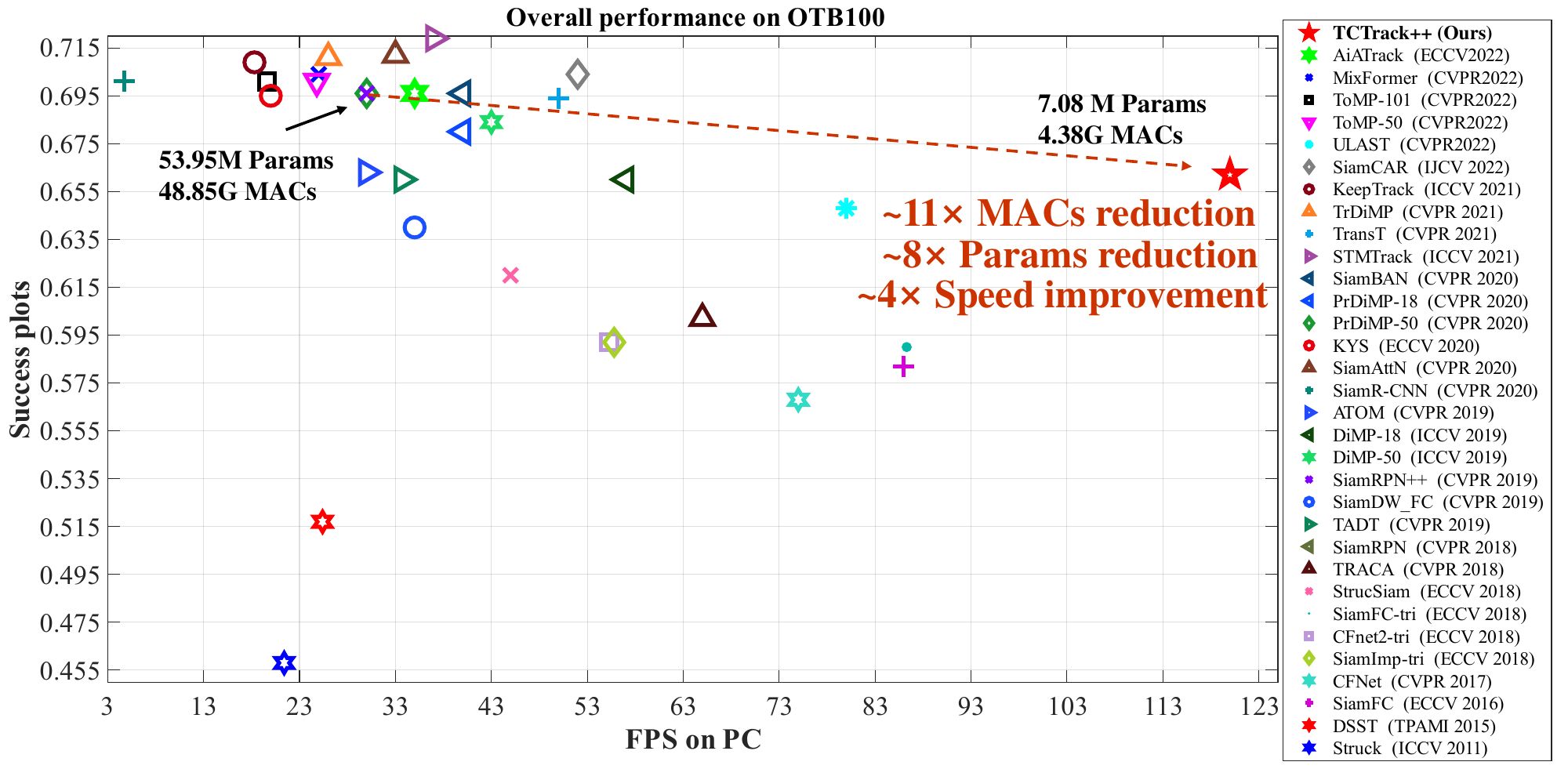}

	\caption{Offline overall performance on OTB100 benchmarks. Our tracker achieves competitive performance against other SOTA trackers. Please note that MACs represent the Multiply–Accumulate Operations. }
	\label{fig:deeper_star1}
	
\end{figure}

	\noindent\textbf{OTB100.} As the most famous tracking benchmark, OTB100 is also used to evaluate offline tracking performance. As shown in Fig.~\ref{fig:deeper_star1}, compared with SiamRPN++~\cite{siamrpn++}, despite 4\% offline performance reduction, our tracker achieves \textbf{11$\times$} multiply-accumulate operations reduction (MACs), \textbf{8$\times$} parameters reduction, and \textbf{4$\times$} speed improvement. Due to the low latency structure and similar tracking performance, our tracker can achieve superior real-world tracking performance as discussed before.

\begin{figure}[t]
	\centering
	\vspace{-10pt}
	\includegraphics[width=1\linewidth]{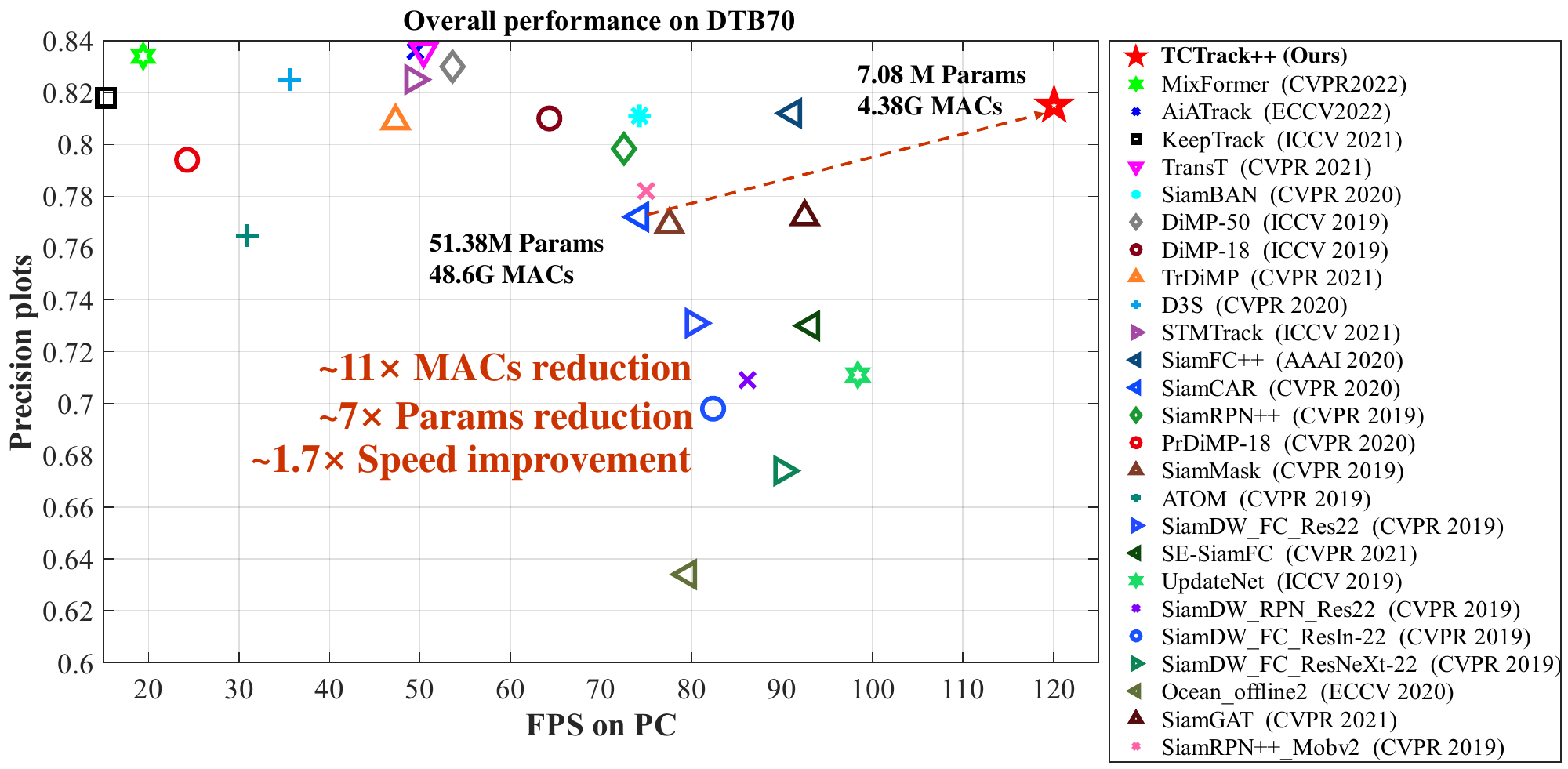}
	
	\caption{Overall performance of all trackers on DTB70 benchmarks. Our tracker achieves competitive performance against other SOTA trackers.}
	
	\label{fig:deeper_star}
\end{figure}

	\noindent\textbf{DTB70.} DTB70~\cite{li2017visual} includes 70 severe motion scenarios in various challenging scenes. The results shown in Fig.~\ref{fig:deeper_star} further validate the competitive tracking performance against other SOTA trackers. Compared with SiamCAR~\cite{siamcar}, we still achieve \textbf{11$\times$} MACs reduction and \textbf{7$\times$} parameters reduction while outperform some deep trackers.

	In conclusion, the comparison with deep SOTA trackers demonstrates the superiority of our tracker in structure and tracking performance, confirming the promising real-world tracking performance on the other side.

	\subsection{Ablation Study}\label{Sec:abla}
	In this subsection, we conduct the ablation studies on five benchmarks including offline evaluations and online evaluations.


\begin{table*}[t]
	\centering
	\scriptsize 
	\caption{Ablation study of different components of the adaptive temporal transformer (AT-Trans) on UAV123~\cite{Mueller2016ECCV}. \texttt{TIF} denotes the temporal information filter in the AT-Trans (Fig.~\ref{fig:work}). \texttt{SF/MF} refer to single-frame (SF) training, \textit{i.e.,} the standard tracking-by-detection training method and our multi-frame (MF) training method. \texttt{CI/RI} refers to convolutional initialization and random initialization for temporal prior knowledge. Query denotes which feature map is used as the query in the adaptive temporal encoder in AT-Trans mentioned in Sec.~\ref{tf}. 
	}
	\vspace{-5pt}
	\renewcommand\tabcolsep{1pt}
	\begin{tabular}{lccc|cccccccc}
		\toprule
		& & & & \multicolumn{2}{c}{Camera Motion} &\multicolumn{2}{c}{Fast motion}&\multicolumn{2}{c}{Partial Occlusion} & \multicolumn{2}{c}{Overall} \\
		\midrule
		Model & Train & Init. & Query & Prec. & Succ. & Prec. & Succ.& Prec. & Succ.& Prec. & Succ. \\
		\midrule
		Transformer & SF & - & $\mathrm{F_{t-1}^m}$ &  $0.750$ & $0.549$&$0.712$ & $0.509$&$0.663$ & $0.458$& $0.750$ & $0.550$ \\
		Transformer+TIF & SF & - & $\mathrm{F_{t-1}^m}$ & $0.767_{2.3\%\uparrow}$ & $0.578_{5.3\%\uparrow}$ & $0.720_{1.1\%\uparrow}$ & $0.525_{3.1\%\uparrow}$ & $0.667_{0.6\%\uparrow}$ & $0.474_{3.5\%\uparrow}$  & $0.765_{2.0\%\uparrow}$ & $0.573_{4.2\%\uparrow}$ \\
		Transformer & MF & CI & $\mathrm{F_{t-1}^m}$ & $0.749_{2.4\%\downarrow}$ & $0.525_{7.6\%\downarrow}$ & $0.719_{2.4\%\downarrow}$ & $0.500_{7.6\%\downarrow}$ &  $0.639_{2.4\%\downarrow}$ & $0.415_{7.6\%\downarrow}$ & $0.732_{2.4\%\downarrow}$ & $0.508_{7.6\%\downarrow}$ \\
		Transformer+TIF & MF & RI & $\mathrm{F_{t-1}^m}$ & $0.779_{3.9\%\uparrow}$  & $0.592_{7.8\%\uparrow}$& $0.766_{7.6\%\uparrow}$  & $0.566_{11.2\%\uparrow}$& $0.670_{1.1\%\uparrow}$  & $0.483_{5.5\%\uparrow}$ & $0.772_{2.9\%\uparrow}$  & $0.586_{6.6\%\uparrow}$\\
		Transformer+TIF & MF & CI & $\mathrm{F_{t}}$ & $0.785_{4.7\%\uparrow}$ & $0.587_{6.9\%\uparrow}$& $0.726_{2.0\%\uparrow}$ & $0.528_{3.7\%\uparrow}$& $0.676_{2.0\%\uparrow}$ & $0.480_{4.8\%\uparrow}$ & $0.771_{2.8\%\uparrow}$ & $0.580_{5.5\%\uparrow}$ \\
		\textbf{Transformer+TIF} & \textbf{MF} & \textbf{CI} & $\bf{F_{t-1}^m}$ & $\bm{0.810_{8.0\%\uparrow}}$ & $\bm{0.615_{12.0\%\uparrow}}$& $\bm{0.793_{11.3\%\uparrow}}$ & $\bm{0.586_{15.1\%\uparrow}}$& $\bm{0.710_{7.1\%\uparrow}}$ & $\bm{0.510_{11.4\%\uparrow}}$& $\bm{0.800_{6.7\%\uparrow}}$ & $\bm{0.604_{9.8\%\uparrow}}$ \\ 
		\bottomrule
	\end{tabular}%
	\label{tab:ablati}%
\end{table*}%

\begin{table*}[t]
	
	\centering
	
	\caption{Ablation studies of different structures of our tracker. \texttt{Curr} represents the curriculum training strategy and \texttt{Length} represents the length of training videos. Note that only the tracker uses curriculum learning, will the video length be variable.}
	\vspace{-10pt}
	\renewcommand\tabcolsep{4pt}
	\resizebox{1.0\linewidth}{!}{
		\begin{tabular}{l|cc |cc| c cc | cc|}
			\toprule
			&\multicolumn{2}{c|}{\multirow{2}*{\textbf{Training}}}&\multicolumn{2}{c|}{\multirow{2}*{\textbf{Temporal Info.}}}&\multicolumn{3}{c|}{\textbf{Offline evaluation}}&\multicolumn{2}{c|}{\textbf{Online evaluation}}\\
			& && &&\multicolumn{1}{c}{\textbf{OTB100}}&\multicolumn{1}{c}{\textbf{UAVTrack112}}&\multicolumn{1}{c|}{\textbf{UAV123}}&\multicolumn{1}{c}{\textbf{OTB100}}&\multicolumn{1}{c|}{\textbf{GOT-10K}}\\
			\midrule
			Different structure&Length&Curr&\multicolumn{1}{c}{Lv. feat}&\multicolumn{1}{c|}{Lv. sim}&\multicolumn{1}{c}{Succ.} &\multicolumn{1}{c}{Succ.}&\multicolumn{1}{c|}{Succ.} &\multicolumn{1}{c}{Succ.} &\multicolumn{1}{c|}{AO.}\\
			
			\midrule
			Spatial backbone+CNN (Baseline)&1&\XSolidBrush&\XSolidBrush&\XSolidBrush& $0.548$  & $0.531$  &$0.509$
			
			&$0.451$ &$0.278$

			\\
			
			Online TAdaConv+AT-Trans&1&\XSolidBrush&\XSolidBrush&\Checkmark& $0.601_{9.7\%\uparrow}$ &$0.584_{10.0\%\uparrow}$ 
			
			&$0.557_{9.4\%\uparrow}$
			
			& $0.455_{0.8\%\uparrow}$&$0.281_{1.1\%\uparrow}$ 
			\\
			Online TAdaConv+AT-Trans  &2&\XSolidBrush&\Checkmark&\Checkmark & $0.615_{12.2\%\uparrow}$ &  $0.594_{11.9\%\uparrow}$ 
			
			&$0.559_{9.8\%\uparrow}$ 
			
			& $0.492_{9.1\%\uparrow}$&$0.274_{1.4\%\downarrow}$ 
			\\
			Online TAdaConv+AT-Trans &3 &\XSolidBrush&\Checkmark&\Checkmark & $0.620_{13.1\%\uparrow}$  &$0.595_{12.1\%\uparrow}$ 
			
			&$0.580_{13.9\%\uparrow}$ 
			
			& $0.502_{11.3\%\uparrow}$ &$0.274_{1.4\%\downarrow}$ 
			\\
			Online TAdaConv+AT-Trans (TCTrack)&4 &\XSolidBrush&\Checkmark &\Checkmark& $0.624_{13.9\%\uparrow}$  &$0.611_{15.1\%\uparrow}$ 
			
			&$0.604_{18.7\%\uparrow}$ 
			
			& $0.518_{14.9\%\uparrow}$ &$0.312_{12.2\%\uparrow}$ \\
			
			ATT-TAdaConv+AT-Trans  &4 &\XSolidBrush&\Checkmark&\Checkmark& $0.642_{17.2\%\uparrow}$ &  $0.627_{18.1\%\uparrow}$ 
			
			&$0.604_{18.7\%\uparrow}$ & 
			
			$0.510_{13.1\%\uparrow}$&$0.333_{19.8\%\uparrow}$ \\
			\textbf{ATT-TAdaConv+AT-Trans (TCTrack++)} &2$\rightarrow$4 &\Checkmark &\Checkmark&\Checkmark& $\bf{0.662_{20.8\%\uparrow}}$ &  $\bf{0.640_{20.5\%\uparrow}}$ 
			
			&$\bf{0.608_{19.4\%\uparrow}}$ 
			
			& $\bf{0.543_{20.4\%\uparrow}}$ &$\bf{0.375_{34.9\%\uparrow}}$  \\

			\bottomrule
	\end{tabular}}
	
	\label{tab:a}%
\end{table*}%

\begin{figure*}[t]
	\centering

	\includegraphics[width=1\textwidth]{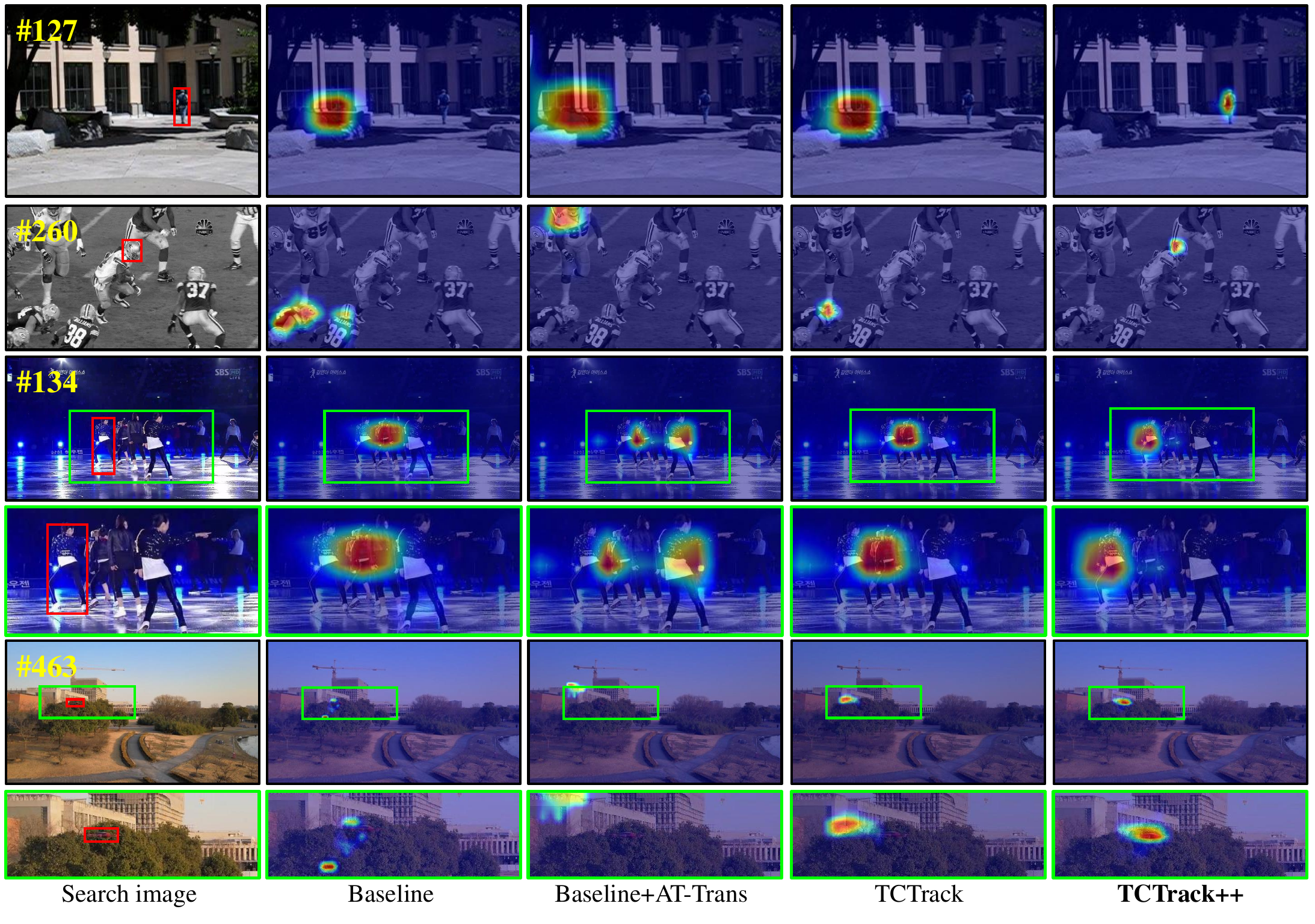}
	\vspace{-25pt}
	\caption{Visualization of trackers with different structure. Red box represent the target object. It shows intuitively the effectiveness of our proposed method and strategy. From the top to the bottom, the sequences are the \textit{Human8}, \textit{Football}, \textit{Skating1} from OTB100 and \textit{UAV4} from UAVTrack112.}
	
	\label{fig:vis}
	
\end{figure*}

	\begin{figure*}[t]
		\centering	
		\includegraphics[width=1\linewidth]{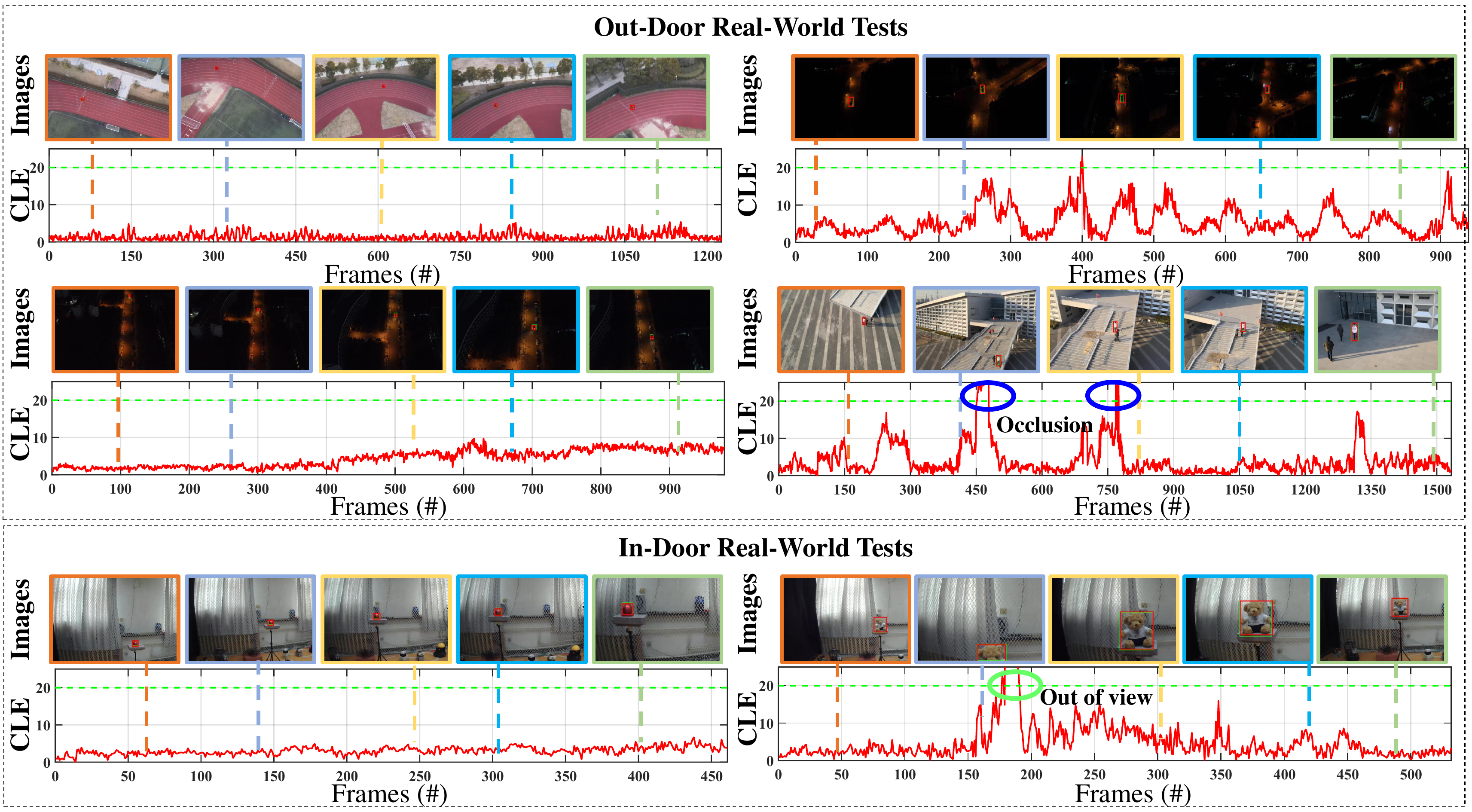}
		\vspace{-20pt}
		\caption{Recording of real-world tests on the embedded platform. The tracking targets are marked with \NoOne{red} while the CLE represents the center location error. To avoid unpredictable disclosure of personally identifiable information, images are processed merely.} 
		
		\label{fig:v4r}
	\end{figure*}
	\noindent\textbf{Clarification of symbol.} To clarify, we denote our proposed transformer architecture without temporal information filter as \texttt{Transformer} in TABLE~\ref{tab:ablati}. In TABLE~\ref{tab:a}, we denote temporal information from feature-level and similarity maps as \texttt{Lv. feat} and \texttt{Lv. sim}. The influence of training strategy (curriculum learning) and length of video are also involved. All different structure trackers are adopted in the same setting except for the studied module.
	
	\noindent\textbf{Analysis on AT-Trans.} We conduct the studies about our AT-Trans based on TCTrack~\cite{cao2022tctrack} shown in TABLE~\ref{tab:ablati}. \textbf{I}) Adding the consecutive temporal knowledge without filtering out the invalid information (third line) will confuse the tracker. Therefore, the tracking performance is impeded significantly. By adding our information filter in the tracking-by-detection framework, our module can also raise its performance by adaptively selecting valid contexts (second line). \textbf{II}) As we discussed before, using the unique information of the tracking object in the first frame to initiate the temporal knowledge is more appropriate than random initiation, especially in occlusion conditions (raising about 6\%). \textbf{III}) We also analyze the effect caused by the different queries. The results prove that refinement based on the current similarity map is more effective and suitable for raising performance, especially in motion scenarios (improved over 10\%). 
	
	Compared with \texttt{Transformer}, there is a significant improvement brought by our temporal knowledge encoded by AT-Trans (\textbf{9.8\%} in overall AUC and \textbf{6.7\%} in overall precision). Specifically, our tracker yields the best performance with an improvement of about \textbf{12.0\%} and \textbf{15.1\%} in handling the motion scenes. In the occlusion conditions, owing to the consecutive temporal contexts, our tracker can  boost the success rate by \textbf{11.4\%}.

	\noindent\textbf{Studies about the ATT-TAdaConv and curriculum learning.} To validate the effectiveness of other modules exhaustively, we extend the ablation studies compared with the conference version. As shown in TABLE~\ref{tab:a}, when the image range of TAdaConv is increasing, the performance of online TAdaConv is raising step by step. Plus the attention mechanism, the accuracy of our temporal backbone, \textit{i.e.}, ATT-TAdaConv, is highly boosted and gains \textbf{17.2\%} improvement compared with baseline structure on OTB100. 
	
	Generally, learning temporal contexts from longer videos is harder than from short ones. Therefore, to accelerate the training process and improve the performance of optimization, we try to adopt the curriculum learning strategy for increasing the length of videos step by step. Based on this strategy, our tracker can learn prior knowledge from easier to harder, raising the flexibility and robustness of our trackers. The results of using curriculum learning strategy (the last line) or not confirm the effectiveness of this training strategy. From the first three lines, we can also find that by benefiting from the temporal contexts via different levels, the accuracy and robustness of the tracker will be raised high.

	We also conduct experiments on ResNet. According to Table~\ref{tab:deeper_avg}, the comparisons show that adopting a deeper backbone will improve offline performance (60.8 vs 66.1) with a significant gap in FPS (10.1 vs 27.1). Meantime, real-world tracking performance is also limited (0.472 vs 0.519). Therefore, to achieve better real-world tracking performance, we adopt AlexNet as our priority.

	Overall, compared with baseline, our proposed methods make a great performance improvement in both offline evaluation and online evaluation. It strongly verifies the effectiveness of our tracker. Especially, our tracker yields the best performance with an improvement of about \textbf{134\%} on GOT-10K. Besides, we also visualize the attention maps of some structures illustrated in Fig.~\ref{fig:vis}. It shows that during tracking processing, owing to our temporal contexts from two levels, our proposed methods can maintain better robustness compared with trackers without them.

	\section{Real-World Tests}\label{sec:Real-world Tests}
	Generally, most existing real-world tracking applications cannot afford the high-end GPU. Therefore, to better validate the real-world tracking performance of our TCTrack++, we deploy it on the resource-limited platform, \textit{i.e.}, NVIDIA Jetson AGX Xavier~\footnote{https://www.nvidia.com/en-us/autonomous-machines/embedded-systems/jetson-agx-xavier/}. Note that to increase the diversity of our tracking scenes, we use our UAV to collect the video and tracking onboard. Our real-world tests cover many challenging scenes including indoor \& outdoor scenes and daytime \& dark time scenes with low resolution, illumination variation, severe occlusion, out-of-view, and camera motion attributes. We use the CLE as the evaluation metric with a threshold (20 pixels).  Additionally, the average speed of TCTrack++ in real-world tests is 27.1FPS with the 53\% and 13.42\% utilization of GPU and CPU on average.

	\subsection{Outdoor Tests}
	
	As shown in Fig.~\ref{fig:v4r}, we choose several challenging scenes in real-world conditions. The stable tracking results of the first scene strongly proves the promising tracking performance in the low-resolution scene. Meantime, our proposed tracker also maintains impressive in very low illumination scenes. Furthermore, TCTrack++ performs impressive accuracy when facing severe occlusion. When the object reappears from the obstacle, our tracker can relocate the target based on temporal contexts.

	\subsection{Indoor Tests}
	
	For enlarging the range of testing scenes, we also involve the indoor scenes. To raise the tracking difficulties, we move the camera frequently and severely and introduce the out-of-view scenes on purpose. Overall, our tracker keeps a superior real-world tracking performance in those challenging scenes.

	\begin{table*}[t]
	\centering
	\caption{Overall performance of our tracker against other SOTA trackers on UAV123. Note that the FPS are tests on NVIDIA AGX Xavier. $\Delta \%$ it represents the decreased ratio of success rate after introducing the latency of hardware.}
	\resizebox{0.99\linewidth}{!}
	{
		\begin{tabular}{lccccccccc}
			\toprule
			Tracker & \bf{TCTrack++} & \bf{TCTrack++}&ToMP101& TrDiMP & SiamCAR & SiamBAN & PrDiMP18 & SiamRPN++ & DiMP50   \\
			\midrule
			Backbone & ResNet50 & AlexNet&ResNet101& ResNet50 & ResNet50 & ResNet50 & ResNet18 & ResNet50 & ResNet50   \\
			\midrule
			FPS. ($\uparrow$) & 10.1 &\bf{27.1}&4.2&6.5&5.8&5.7&11.2&5.6&9.8 \\
			Offline Succ. ($\uparrow$) & 0.661& 0.608 & 0.669  & \bf{0.675}  & 0.614  & 0.631  & 0.653 & 0.613  & 0.653   \\
			Online Succ. ($\uparrow$) & 0.472& \bf{0.519} & 0.339  & 0.356  & 0.358  & 0.365  & 0.490 & 0.359  & 0.447   \\
			\midrule
			$\Delta \%$ ($\downarrow$) & 28.6& \bf{14.6} & 49.3  & 47.3  & 41.7  & 42.2  & 25.0 & 41.4  & 31.5   \\
			\bottomrule
	\end{tabular}}
	\vspace{-8pt}
	\label{tab:deeper_avg}%
\end{table*}%

	\section{Conclusion and Discussion}\label{sec:conclusion}
	In this work, we propose a comprehensive and efficient tracking framework to exploit temporal contexts for real-world visual tracking. Specifically, AT-Trans and ATT-TAdaConv are proposed for exhaustively exploring temporal contexts via two levels. Besides, attributing to our online updating strategy, unnecessary operations and memory loading are avoided. Additionally, compared with our previous version, we also introduce curriculum learning into visual tracking to further boost the tracking performance. Exhaustive online and offline experiments on eight benchmarks strongly demonstrate superior real-world tracking performance and impressive efficiency. Besides, to validate the effectiveness of our tracker, ablation studies are conducted. We believe that our framework can inspire future research in visual tracking.
	
	

	
	%

	\appendices
	
	 \section{Future Works and Research Prospects}
	In this research paper, we have developed a promising temporal-based tracker for real-world tracking. Through experiments conducted on 8 benchmarks, we have already demonstrated the promising speed and accuracy of our tracker. In this section, we will further validate the effectiveness and robustness of our proposed models in deeper backbone, \textit{i.e.}, ResNet. The comparison among other trackers is shown in TABLE~\ref{tab:deeper_avg}. 
		Our tracker demonstrates impressive online/offline performances and frames per second (FPS) compared to other trackers utilizing ResNet50. Considering the current hardware performance and online applications, we still choose AlexNet as our priority in this paper. In our future work, we will focus on resolving the conflict between deeper backbone architectures and latency, aiming to find a balance that maximizes performance while minimizing latency.

	

	\ifCLASSOPTIONcompsoc
	\section*{Acknowledgments}
	\else
	\section*{Acknowledgment}
	\fi
	
	This study is supported by the National Natural Science
	Foundation of China (No.62173249), the Natural Science
	Foundation of Shanghai (No.20ZR1460100), and the Ministry of Education, Singapore, under its MOE AcRF Tier 2 (MOE-T2EP20221- 0012), NTU NAP, and under the RIE2020 Industry Alignment Fund – Industry Collaboration Projects (IAF-ICP) Funding Initiative, as well as cash and in-kind contribution from the industry partner(s). $\hfill\blacksquare$
	
	\ifCLASSOPTIONcaptionsoff
	\newpage
	\fi

	
	
	%
	%
	%
	\normalem
	\bibliographystyle{IEEEtran}
	\bibliography{journal}
	%
	
	\begin{IEEEbiography}[{\includegraphics[width=1in,height=1.25in,clip,keepaspectratio]{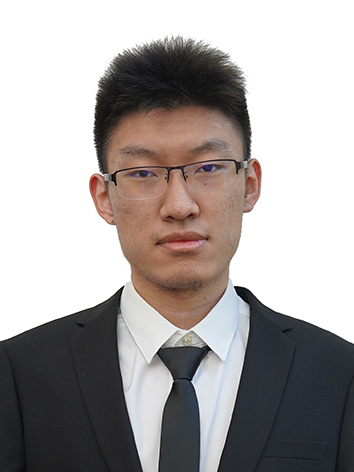}}]{Ziang Cao}
		received the B.Eng. degree at the School of Automotive Studies in Tongji University, Shanghai, China. His research interests include computer vision, deep learning and 3D generation.
	\end{IEEEbiography}
	
	\begin{IEEEbiography}[{\includegraphics[width=1in,height=1.25in,clip,keepaspectratio]{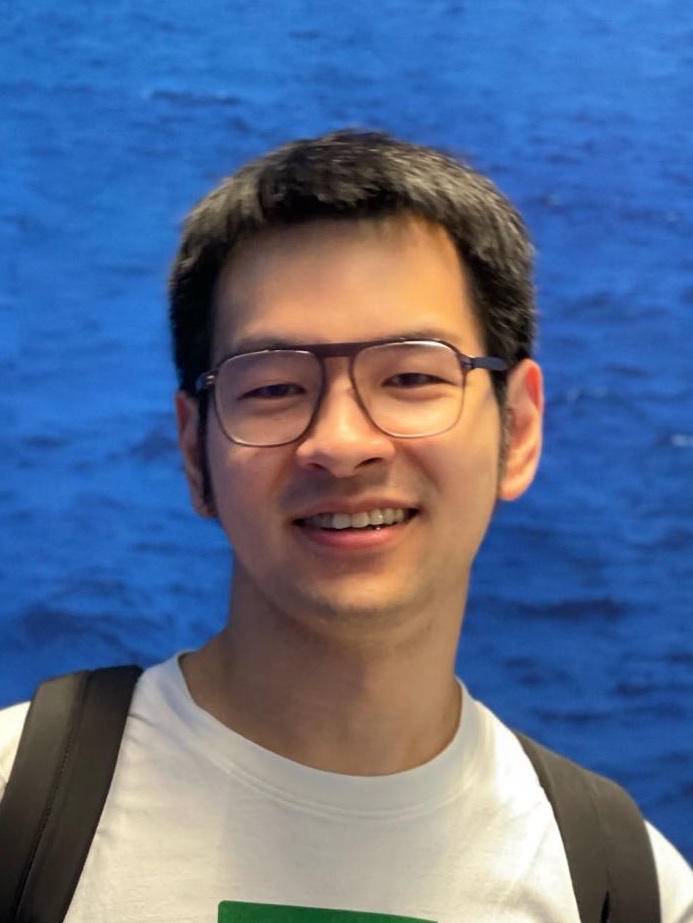}}]{Ziyuan Huang}
		received the B. Eng. degree in Vehicle Engineering from Tongji University in 2019, and is currently a Ph.D. candidate in Advanced Robotics Centre at National University of Singapore, supervised by Professor Marcelo Ang. His main research interests are on video understanding, including action recognition and localization, video representation learning, multi-modal learning, and video-based scene understanding.
	\end{IEEEbiography}
	
	\begin{IEEEbiography}[{\includegraphics[width=1in,height=1.25in,clip,keepaspectratio]{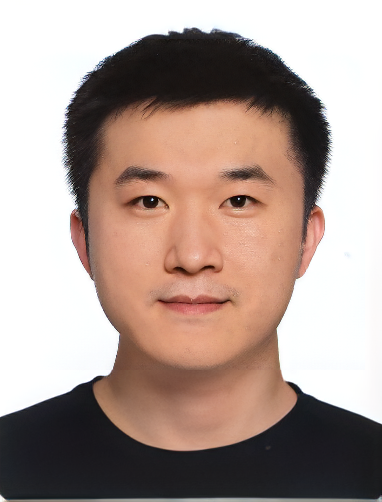}}]{Liang Pan}
		received the PhD degree in Mechanical Engineering from National University of Singapore (NUS) in 2019.
		He is currently a Research Fellow at S-Lab, Nanyang Technological University (NTU).  
		Previously, He is a Research Fellow at the Advanced Robotics Centre from National University of Singapore.
		His research interests include computer vision and
		3D point cloud, with focus on shape analysis, deep learning, and 3D human. 
		He also serves as a reviewer for top computer vision and robotics conferences, such as CVPR, ICCV, IROS and ICRA.
	\end{IEEEbiography}
	
	
	
	\begin{IEEEbiography}[{\includegraphics[width=1in,height=1.25in,clip,keepaspectratio]{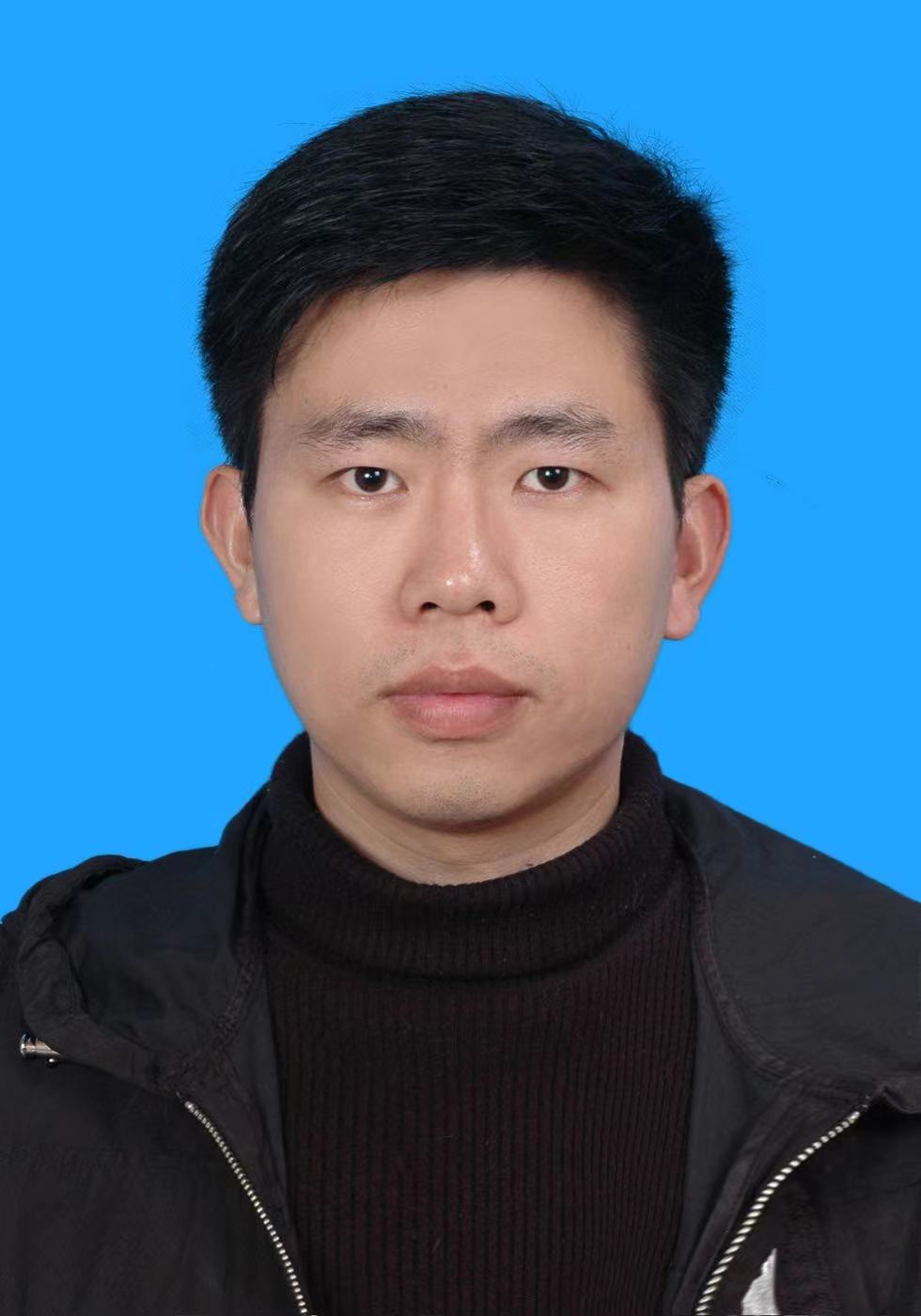}}]{Shiwei Zhang}
		received his Ph.D. degree from the School of Artificial Intelligence and Automation, Huazhong University of Science and Technology in 2019. 
		He is recently a researcher at Alibaba Group.
		His research interests include action detection, action recognition, and multi-modal video understanding.
	\end{IEEEbiography}
	
	\begin{IEEEbiography}[{\includegraphics[width=1in,height=1.25in,clip,keepaspectratio]{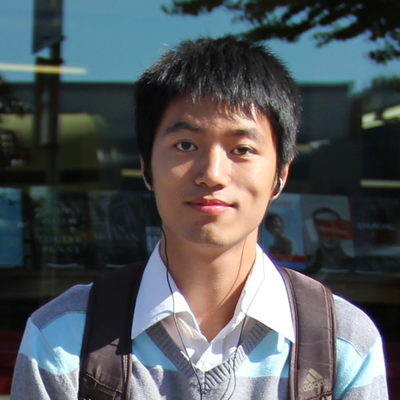}}]{Ziwei Liu}
		Ziwei Liu is currently an Assistant Professor at Nanyang Technological University (NTU). Previously, he was a senior research fellow at the Chinese University of Hong Kong and a postdoctoral researcher at the University of California, Berkeley. Ziwei received his Ph.D. from the Chinese University of Hong Kong in 2017. His research revolves around computer vision/graphics, machine learning, and robotics. He has published extensively on top-tier conferences and journals in relevant fields, including CVPR, ICCV, ECCV, NeurIPS, IROS, SIGGRAPH, TOG, and TPAMI. He is the recipient of the Microsoft Young Fellowship, Hong Kong PhD Fellowship, ICCV Young Researcher Award, and HKSTP best paper award.
		
	\end{IEEEbiography}

	\begin{IEEEbiography}[{\includegraphics[width=1in,height=1.25in,clip,keepaspectratio]{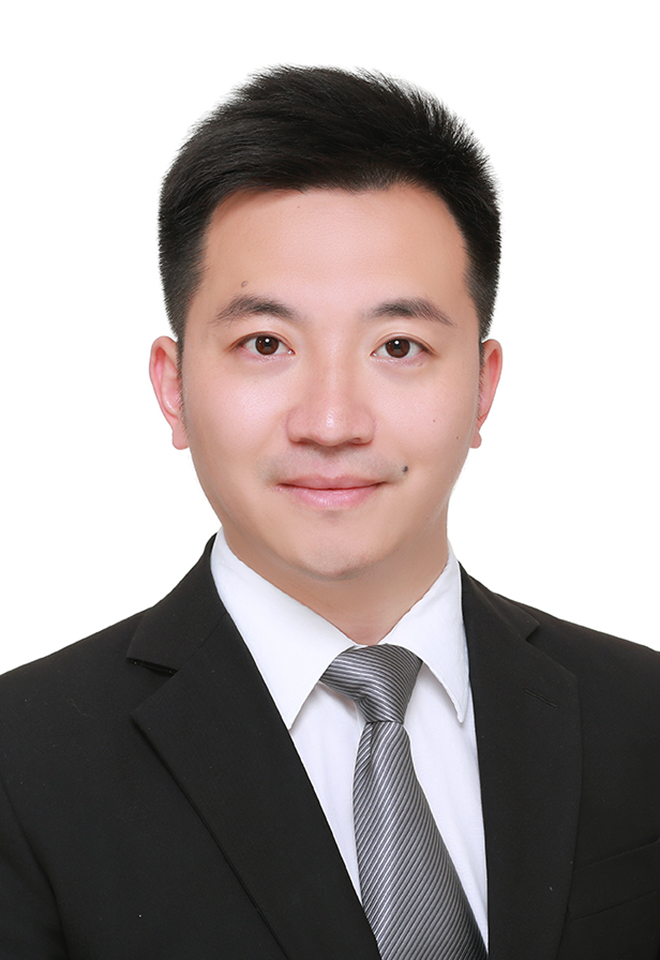}}]{Changhong Fu}
	(Member, IEEE) received the Ph.D. degree in robotics and automation from the Computer Vision and Aerial Robotics (CVAR) Laboratory, Technical University of Madrid, Madrid, Spain, in 2015. During his Ph.D., he held two research positions at Arizona State University, Tempe, AZ, USA, and Nanyang Technological University (NTU), Singapore. After receiving his Ph.D., he worked at NTU as a Post-Doctoral Research Fellow. He is currently an Associate Professor with the School of Mechanical Engineering, Tongji University, Shanghai, China, and leading more than five projects related to the vision for unmanned systems (US). In addition, he has published more than 100 journal and conference papers (including the IEEE GRS Magazine, AI Review, IEEE TCSVT, IEEE TMM, IEEE/ASME TMECH, IEEE TIE, IEEE TII, IEEE TGRS, IEEE TMC, IEEE TITS, IEEE RA Letters, IEEE CVPR, IEEE ICCV, IEEE ICRA, IEEE IROS) related to the intelligent vision and control for US. His research areas are intelligent vision and control for US in a complex environment.
	\end{IEEEbiography}


	
	

\end{document}